\documentclass[final,3p]{elsarticle}
\usepackage{geometry}
\usepackage[colorlinks,linkcolor=blue,anchorcolor=blue,citecolor=blue]{hyperref}
\usepackage{amssymb}
\usepackage{mathrsfs}
\usepackage{amsmath}
\usepackage{textcomp}
\usepackage{bm}
\usepackage{booktabs}
\usepackage{multirow}
\usepackage{graphicx}
\usepackage{epstopdf}
\usepackage{caption}

\def\equationautorefname~#1\null{(#1)\null}
\usepackage{array}
\usepackage{url}
\usepackage{upgreek}
\usepackage{tikz}
\graphicspath{ {./} }
\usepackage{subcaption}
\usepackage[ruled,linesnumbered]{algorithm2e}
\SetAlFnt{\footnotesize}
\SetAlCapFnt{\footnotesize}
\SetAlCapNameFnt{\footnotesize}
\usepackage[norule,hang,flushmargin]{footmisc}
\usepackage{lipsum}

\journal{Computer Methods in Applied Mechanics and Engineering}

\begin{document}

\begin{frontmatter}

\title{A clustering adaptive Gaussian process regression method: response patterns based real-time prediction for nonlinear solid mechanics problems}

\author[add1,add2]{{Ming-Jian Li}}
\author[add1,add2]{{Yanping Lian}\corref{mycorrespondingauthor} }
\cortext[mycorrespondingauthor]{Corresponding author: yanping.lian@bit.edu.cn (Yanping Lian)}
\author[add1]{{Zhanshan Cheng}}
\author[add1]{{Lehui Li}}
\author[add1]{{Zhidong Wang}}
\author[add1,add2]{{Ruxin Gao}}
\author[add1,add2]{{Daining Fang}}
\address[add1]{Institute of Advanced Structure Technology, Beijing Institute of Technology, Beijing 100081, China}
\address[add2]{Beijing Key Laboratory of Lightweight Multi-functional Composite Materials and Structures, Beijing 100081, China}

\begin{abstract}

Numerical simulation is a powerful approach to study nonlinear solid mechanics problems. 
However, mesh-based or particle-based numerical methods suffer from the common shortcoming of being time-consuming, particularly for complex problems with real-time analysis requirements. 
This study presents a clustering adaptive Gaussian process regression (CAG) method aiming for real-time prediction for nonlinear structural responses in solid mechanics. 
It is a data-driven machine learning method featuring a small sample size, high accuracy, and high efficiency, leveraging nonlinear structural response patterns.  
Similar to the traditional Gaussian process regression (GPR) method, it operates in offline and online stages. 
In the offline stage, an adaptive sample generation technique is introduced to cluster datasets into distinct patterns for demand-driven sample allocation. This ensures comprehensive coverage of the critical samples for the solution space of interest.
In the online stage, following the divide-and-conquer strategy, a pre-prediction classification categorizes problems into predefined patterns sequentially predicted by the trained multi-pattern Gaussian process regressor.
In addition, dimension reduction and restoration techniques are employed in the proposed method to enhance its efficiency.
A set of problems involving material, geometric, and boundary condition nonlinearities is presented to demonstrate the CAG method's predictive abilities. 
The proposed method can offer predictions within a second and attain high precision with only about 20 samples within the context of this study, outperforming the traditional GPR using uniformly distributed samples for error reductions ranging from 1 to 3 orders of magnitude. 
The CAG method is expected to offer a powerful tool for real-time prediction of nonlinear solid mechanical problems and shed light on the complex nonlinear structural response pattern.

\end{abstract}


\end{frontmatter}

\section{Introduction}
\label{secIntro}

Solid mechanics confronts a diverse range of complex material, geometric, and boundary condition nonlinearities, often necessitating the utilization of numerical simulation algorithms. However, it is time-consuming to conduct high-fidelity numerical simulations utilizing the mesh or particle-based numerical methods for large-scale problems as the complexity rises, especially in scenarios demanding immediate assessment, such as online structural damage evaluation during high-speed impacts, seismic loads, and other severe serving conditions. Data-driven methods offer a promising path forward, revolutionizing prediction speed, and are believed to be a new scientific paradigm following the computational sciences \citep{szalay20062020,hey2009fourth}.

For the time being, data-driven methods can be broadly categorized into two types, large-sample based methods and small-sample based methods. The former demands extensive data for model training to ensure prediction accuracy and generalization ability.  Neural networks evolving from the perceptron algorithm \citep{rosenblatt1957perceptron} are the most prevalent models in large-sample based methods. They consist of interconnected neurons in a hierarchical structure, mirroring the complexity of the human brain. Their advantages lie in nonlinear activation functions and the universal approximation theorem \citep{cybenko1989approximation,hornik1989multilayer}, allowing them to excel in addressing nonlinearity in solid mechanics problems \citep{liu2022mechanistically,liang2023synergistic,bai2024robust,ghnatios2024new,goodbrake2024neural}. Fuhg et al. \cite{fuhg2022learning} proposed a tensor basis neural network for hyperelastic anisotropy prediction, with three hidden layers of 30 neurons and 2,500 samples. Masi et al. \cite{masi2021thermodynamics} introduced a thermodynamics-driven artificial neural network (ANN) for elastoplastic material behavior, using several hundred neurons and 6,000 samples (3,000 for training). Chen et al. \cite{chen2023full} put forward a model comprising two sequential convolutional neural networks (CNN) for stress and crack analysis in composites, using 7,200 samples with 87\% dedicated to training. Li et al. \cite{li2022equilibrium} presented an equilibrium-based convolutional neural network (ECNN) to model the constitutive behavior of hyperelastic materials. This model employed 21,063 parameters and was trained on 16,848 samples obtained from finite element simulations. Worthington et al. \cite{worthington2023crack} employed feed-forward artificial neural networks with a multilayer perceptron architecture to predict crack growth. The model used a single hidden layer of 100 hidden nodes and was trained on a dataset comprising 49,050 observations, which were divided into training, validation, and testing datasets in a 3:1:1 ratio.
As obscuring the inner workings of their predictions, these models are viewed as black boxes and necessitate complex architectures and numerous parameters to achieve high accuracy at the expanse of a substantial amount of training dataset \citep{shalev2014understanding}.
Consequently, these models have limited interpretability and inevitable data-availability issues, both of which are not favorable to complex nonlinear solid mechanics problems. 

The other type of data-driven methods may achieve the required performance with a limited dataset, avoiding the time-consuming and costly data preparation process. 
For examples, the ridge regression \citep{hoerl1970ridge,zhang2022machine}, least absolute shrinkage and selection operator (LASSO) regression \citep{tibshirani1996regression,lee2021model,wang2022establish}, support vector regression \citep{smola2004tutorial,bonifacio2019application}, and Gaussian process regression (GPR) \citep{rasmussen2003gaussian,berger2013statistical} have proven to be effective with limited samples. 
In these small-sample based methods, the prior knowledge or specific techniques are usually employed to decrease the complexity of models, mitigating overfitting in the small dataset.
Among them, GPR is a representative method, employing the Gaussian process as a prior distribution and the Bayesian inference to obtain a posterior distribution. Compared to neural networks, GPR features simplicity of implementation, adaptive hyperparameter acquisition, better interpretability, and flexible non-parametric inference. Moreover, GPR goes beyond mere mean predictions, offering variances and confidence intervals, and thus affords the probabilistic assessment of the prediction's trustworthiness. 
Guo et al. \cite{guo2018reduced} presented a GPR algorithm with proper orthogonal decomposition for dimension reduction to predict structural behaviors involving finite deformations and elastoplastic constitutive relations. 
Wang et al. \cite{wang2021metamodeling} documented the capability of the GPR with the singular value decomposition (SVD) for data compression to build metamodels of a nonlinear constitutive model for viscoelastic hydrogels.
Ozturk et al. \cite{ozturk2021uncertainty} employed GPR with a crystal plasticity finite element model for an uncertainty quantification framework aiming to determine a constitutive model for polycrystalline Ti alloys. 
Most recently, He et al. \cite{he2023dual} introduced a dual order-reduced Gaussian process emulator, incorporating principal component analysis (PCA) and Gaussian process regression to forecast stress intensity factors and cracking surfaces for crack problems. 
Athanasiou et al. \cite{athanasiou2023integrated} applied GPR integrated with in-situ experiments and high-fidelity simulations in characterizing fracture instability for indentation pillar-splitting problems. 
Nevertheless, the small sample size brings the challenge of deciphering the suitable training samples and avoiding interference from low-relevant samples via the kernel function, which is employed to determine the pivotal covariance matrix in the GPR. In particular, the complex nonlinear problems in solid mechanics often exhibit distinguishable patterns in physical fields driven by different conditions, such as variations in vibration response, nonlinear deformation, and crack propagation path, just to name a few. Sparse samples may miss capturing certain patterns, resulting in significant sampling bias and compromising prediction accuracy. Data stemming from an infrequent pattern may even be misidentified as noise. In solid mechanics problems, these issues associated with the GPR still remain under investigated..

In this study, we propose a clustering adaptive Gaussian process regression (CAG) method to resolve the abovementioned challenges. 
In the proposed method, the demand-driven sample allocation is achieved offline, and the divide-and-conquer strategy is implemented online, where a dimensionality reduction technique is utilized to improve efficiency. 
In the offline stage, it first adaptively generates training datasets clustered into distinct patterns with an unsupervised learning technique and trains the GPR for each pattern to obtain the multi-pattern Gaussian process regressor. In the online stage, it classifies the specific problem into one of these patterns and finally makes the prediction accordingly. 
As a result of our method, the impact of less pertinent datasets is significantly reduced, leading to highly precise predictions even with a small number of samples. 
A series of numerical cases successfully predicts solid mechanics problems characterized by nonlinearities and multiple patterns, demonstrating the accuracy and efficiency of the CAG method. 

The remainder of this paper is organized as follows. In Section \ref{secMethod}, we present a comprehensive introduction to the proposed CAG method, including its methodology, the offline and online stages, and the implementation procedure. 
Section \ref{secCase} carries out rigorous assessments to evaluate the performance of the CAG method in various challenging scenarios. Notably, the complex nonlinear examples of hyperelastic cylinder under axial stretching and compression, elastoplastic T-shaped structure under extreme twisting, circular ring under high-speed impact, and dynamical crack branching of a brittle plate are studied, showcasing the accuracy, efficiency, and robustness of the proposed method. 
Finally, in Section \ref {secCon}, we draw several significant conclusions that underscore the importance of the CAG method in solid mechanics.

\section{Clustering adaptive Gaussian process regression method}
\label{secMethod}

\subsection{Methodology}
The proposed clustering adaptive Gaussian process regression (CAG) method is 
for the complex nonlinear solid mechanics problem with multi-response patterns. 
It consists of offline and online stages, embodying the demand-driven sample allocation and the divide-and-conquer strategy, respectively. 
In the offline stage, a clustering-enhanced adaptive sample generation scheme is introduced and implemented. This scheme, utilizing unsupervised learning, categorizes samples into distinct patterns and generates new training samples accordingly, ensuring comprehensive coverage of all patterns. This efficient approach fulfills the demand-driven sample allocation. Subsequently, the classified training samples are used to train a supervised learning model for the individual pattern, forming the multi-pattern Gaussian process regressor, which is integrated with a dimensionality reduction technique for further efficiency improvement. During the online stage, a pre-prediction classification method is first employed to categorize the given case into one of the predefined patterns, leveraging the multi-pattern training samples obtained during the offline stage. Subsequently, the trained multi-pattern Gaussian process regressor makes prediction based on the categorized pattern, effectively reducing interference from unrelated training samples, fulfilling the divide-and-conquer strategy. The details of the CAG method are introduced in subsequent sections.

\subsection{Clustering-enhanced adaptive sample generation}
\label{subseccaasg}
To construct the labeled dataset according to the structural response patterns of the problem of interest, we propose the clustering-enhanced adaptive sample generation scheme. 
It consists of three parts: initial sample generation, samples clustering based on structural response patterns, and adaptive sample addition, which are detailed as follows. An important characteristic to note is that the inadequacy of the initial samples for prediction necessitates multiple iterations of clustering and adaptive sample generation. 

\subsubsection{Initial sample generation}
For a given problem, an initial matrix of controlling parameters $\chi$ is firstly established, with a small initial sample size of $M^0$ (typically less than ten). It reads:
\begin{equation}
\boldsymbol X^0 = \left[ \chi_{1}^0 , \; \chi_{2}^0 , \; \cdots , \;\chi_{M^0}^0 \right]^{\text T} \in \mathbb{R}^{M^0}
\label{eq_X0}
\end{equation}
where $\mathbb{R}$ denotes the real number space with superscripts indicating the matrix's dimensions. Let the initial controlling parameters uniformly span the given parameter range $\left[\chi_{\text{min}}, \chi_{\text{max}} \right]$, and then the 
$i$-th sample, $\chi_i^0$, is determined as:
\begin{equation}
\chi_i^0  = \chi_\text{min} + (i-1)\frac{\chi_\text{max} - \chi_\text{min}}{M^0-1} ,\quad i \in [1,M^0]
\label{eq_xii}
\end{equation}

For each loading case characterized by $\chi$, the corresponding structural responses of interest, denoting by the physical field $\boldsymbol{\eta}$, are obtained from high-fidelity numerical simulations as follows.
\begin{equation}
\boldsymbol{\eta}_i^0  =\mathcal {S}   \left( \chi_i^0  \right) ,\quad i\in[1,M^0]
\label{eq_simulation}
\end{equation}
where $\mathcal {S}$ represents the high-fidelity numerical simulation solver. The physical field meets $\boldsymbol{\eta}_{i}\in \mathbb{R}^{N}$ for the $i$-th sample, with $N$ being the dimension of the field and determined by the count of spatial points or temporal intervals. Therefore, the physical field matrix is formulated as: 
\begin{equation}
\boldsymbol{Y}^0  =\left[ \boldsymbol{\eta}_{1}^0, \; \boldsymbol{\eta}_{2}^0,\;  \ldots,\;  \boldsymbol{\eta}_{M_0}^0 \right]
\label{eq_y0}
\end{equation}
with the matrix meeting $\boldsymbol{Y}^0 \in \mathbb{R}^{N \times M^0  }$.
The obtained results from the $M^0$ samples and their controlling parameters serve as the initial labelled input-output pairs, denoted as $\mathbb D ( \boldsymbol{X}^0 , \boldsymbol{Y}^0 )$, which are further studied in the following sections.

\subsubsection{Samples clustering based on structural response patterns}

For the $n$-th iteration, the input dataset $\mathbb D( \boldsymbol{X}^n , \boldsymbol{Y}^n )$ comprises $M^n$ samples. 
In this study, the $K$-means clustering algorithm is employed to categorize these samples into $K$ distinct patterns according to their structural responses' similarities, as follows.

\begin{enumerate}[(1)]
    \item Generate a matrix $\boldsymbol{C}^n \in \mathbb{R}^{N \times K} $ with random numbers as the initial cluster centroids of the physical fields
    \begin{equation}
    \boldsymbol{C}^n  =\left[
    \boldsymbol{\overline{\eta}}_1^n  ,\;
    \boldsymbol{\overline{\eta}}_2^n  ,\;
    \ldots , \;
    \boldsymbol{\overline{\eta}}_K^n
    \right]
    \label{eq_initcent}
    \end{equation}
    where $\boldsymbol{\overline{\eta}}_k^n \in \mathbb{R}^{N} $ denotes the centroid array of the $k$-th cluster. 

    \item Assess the similarity by computing the Euclidean distances $\boldsymbol{\iota}_i^n$ from the physical fields to the cluster centroids: 
    \begin{equation}
    \boldsymbol{\iota}_i^n = [\iota_{i1}^n, \iota_{i2}^n, \ldots, \iota_{iK}^n], \quad  \iota_{ik}^n = \| \boldsymbol{\eta}_{i}^n - \boldsymbol{\overline{\eta}}_k^n \| ,  \quad i \in [1,M^n],\quad k\in [1, K]
    \label{eq_edist}
    \end{equation}
    where $\iota_{ik}^n$ denotes the Euclidean distance from the physical field of the $i$-th sample to the $k$-th cluster centroid. 

    \item Categorize all physical fields into clusters by the nearest centroid principle:
    \begin{equation}
     \boldsymbol{\eta}^n_i \in \boldsymbol{\hat{Y}}_k^n  , \quad \text{if} \quad \iota_{ik} = \text{min} (\boldsymbol{\iota}_i), \quad i\in[1, M^n], \quad k\in [1, K]
    \label{eq_put}
    \end{equation}
     where $\boldsymbol{\hat{Y}}_k^n$ is the clustered physical fields with the caret symbol denoting clusters. 
    
    \item Re-evaluate the centroids of clusters:
    \begin{equation}
    \boldsymbol{\overline{\eta}}_k^n  =  \frac{1}{\hat{M}_{k}^n} \textstyle \sum_{\boldsymbol{\eta}_i^n \in \boldsymbol{\hat{Y}}_k^n } \boldsymbol{\eta}_i^n ,
    \quad i\in[1, M^n], \quad k\in [1, K]
    \label{eq_updatecent}
    \end{equation}
   where $\hat{M}_{k}^n$ is the number of samples in the current $k$-th cluster.
    
    \item Repeat steps (2) to (4) consecutively until the change in cluster centroids, denoted by $\boldsymbol{C}^n$, becomes less than the predefined tolerance.
\end{enumerate}

Upon completion of the clustering steps, the physical field matrix $\boldsymbol{Y}^n$ is divided into $K$ non-overlapping clusters, expressed as $\boldsymbol{\hat{Y}}_k^n$. Each sample in the physical field matrix is allocated to the most corresponding cluster, according to the pattern of structural responses, and the selection of $K$ will be discussed later in Section \ref{subsectwist}. 

\subsubsection{Adaptive sample addition}

Since the samples in clusters are used to train the respective regressions of the CAG, it is imperative that each cluster must contain an adequate number of samples to maintain the respective prediction accuracy. With $Q_{\text {min}}$ denoting the minimum number of samples required for training each pattern's model, the number of samples in each cluster, denoted by $\hat{M}_{k}^n$, must satisfy the following equation:
\begin{equation}
 \forall \hat{M}_{k}^n \geqslant Q_{\text {min}} , \quad k\in [1, K]
 \label{eq_Msample}
\end{equation}
Otherwise, new samples must be added. 
It is best to add the new samples in such a way that the boundaries of the clusters become clearer. Therefore, the new sample is determined as follows.
\begin{equation}
  \chi_\dag^{n+1}  = \frac{1}{2} \left( \chi_i^n+ \chi_{i+1}^n  \right), \quad 
  \text{if} \;\; {\boldsymbol \eta}_{i}^n \in \boldsymbol{\hat{Y}}_k^n  \;\; \text {and} \;\; {\boldsymbol \eta}_{i+1}^n \notin \boldsymbol{\hat{Y}}_k^n
  \quad i\in[1, M^n], \quad k\in [1, K]
  \label{eq_xiadd}
\end{equation}
where $\chi_\dag^{n+1}$ denotes a new controlling parameter close to cluster boundaries. Meanwhile, the high-fidelity numerical simulation is conducted to obtain the physical field $\boldsymbol{\eta}_\dag^{n+1}$ as:
\begin{equation}
\boldsymbol{\eta}_\dag^{n+1}  =\mathcal {S}   \left( \chi_\dag^{n+1}  \right) 
\label{eq_simulationnew}
\end{equation}

The dataset is iteratively expanded by inserting new points $\chi_\dag^{n+1}$ and $\boldsymbol{\eta}_\dag^{n+1}$ into $\mathbb D (\boldsymbol{X}^n , \boldsymbol{Y}^n )$ to form $\mathbb D ( \boldsymbol{X}^{n+1} , \boldsymbol{Y}^{n+1} )$. Such a process of clustering and sample augmentation is iterated until the dataset that meets the criterion in Eq.~\autoref{eq_Msample} is obtained. The final dataset is denoted by
\begin{equation}
\mathbb D ( \boldsymbol{X} , \boldsymbol{Y} )  = \mathbb D ( \boldsymbol{\hat{X}}_1 , \boldsymbol{\hat{Y}}_1 ) \cup \mathbb D ( \boldsymbol{\hat{X}}_2 , \boldsymbol{\hat{Y}}_2 ) \cup \ldots \cup \mathbb D ( \boldsymbol{\hat{X}}_K , \boldsymbol{\hat{Y}}_K )
\label{eq_finalXY}
\end{equation}
where $\mathbb D ( \boldsymbol{X} , \boldsymbol{Y} )$ is divided into $K$ clusters, and the final sample size $M$ satisfies:
\begin{equation}
 M = \sum_{k=1}^K \hat{M}_k
 \label{eq_finalM}
\end{equation}
The process of the clustering-enhanced adaptive sample generation method is concluded in \autoref{algocluster} as follows. 
\begin{algorithm}
\caption{The clustering-enhanced adaptive sample generation method}\label{algocluster}
\KwIn{ $\chi_\text{min}$, $\chi_\text{max}$, $M^0$, $K$, $Q_{\text {min}}$}
\KwOut{$\mathbb D ( \boldsymbol{X} , \boldsymbol{Y} )$}
Generate initial dataset $ \mathbb D ( \boldsymbol{X}^0 , \boldsymbol{Y}^0 ) $ with $M^0$ samples by Eqs.~\autoref{eq_X0}, \autoref{eq_xii}, \autoref{eq_simulation} and \autoref{eq_y0}\;
\While{convergence condition Eq.~\autoref{eq_Msample} is not met}
{
    Initialize $K$-means cluster centroids by Eq.~\autoref{eq_initcent}\;
    \While{cluster centroids $\boldsymbol{C}^n$ is not converged}
    {
    Calculate Euclidean distance between physical fields and cluster centroids by Eq.~\autoref{eq_edist} \;
    Categorize samples into clusters by Eq.~\autoref{eq_put} \;
    Update cluster centroids by Eq.~\autoref{eq_updatecent} \;
    }
    Generate new samples at cluster boundaries by Eqs.~\autoref{eq_xiadd} and \autoref{eq_simulationnew} \;
    Update the dataset $\mathbb D ( \boldsymbol{X}^n , \boldsymbol{Y}^n)$ with new samples to form $\mathbb D( \boldsymbol{X}^{n+1} , \boldsymbol{Y}^{n+1})$\;
}
Obtain the final dataset $\mathbb D ( \boldsymbol{X} , \boldsymbol{Y} )$ in Eq.~\autoref{eq_finalXY}\;  
\end{algorithm}

\subsection{Multi-pattern Gaussian process regressor}

Upon the obtained labelled dataset $\mathbb D ( \boldsymbol{X} , \boldsymbol{Y} )$ and the corresponding clustered subsets, a set of GPRs is trained individually on each cluster. The combination of the trained $K$ GPRs is termed as multi-pattern Gaussian process regressor, aiming to address distinct structural behavior patterns separately. 
Moreover, the physical field of the $k$-th cluster, represented by $\boldsymbol{\hat{Y}}_k \in \mathbb{R}^{N \times \hat{M}_k}$, exhibits a notable imbalance in its dimensions. While the number of samples within the cluster $\hat{M}_{k}$ is typically small (less than 20), the dimension $N$ determined by spatial and temporal intervals of high-fidelity solver $\mathcal{S}$ can be quite large, potentially exceeding hundreds of thousands or more, resulting in time-consuming for the matrix calculations in the GPR. Therefore, dimensionality reduction is conducted to the original dataset.  

\subsubsection{Dimensionality reduction of physical fields}
\label{subsecdr}

In this study, we employed the principal component analysis (PCA) to conduct the dimensionality reduction, aiming to improve the efficiency of matrix calculations of the multi-pattern Gaussian process regressor. Based on the PCA, a reduced-order model denoted as $\boldsymbol{\hat{Y}}_k^\downarrow \in \mathbb{R}^{R \times \hat{M}_k}$ derived from $\boldsymbol{\hat{Y}}_k$ reads:

\begin{equation}
\boldsymbol{\hat{Y}}_k^\downarrow = \mathcal{R}^\downarrow \left( \boldsymbol{\hat{Y}}_k \right)
\label{eq_reduce}
\end{equation}
where $\mathcal{R}^\downarrow$ represents the dimensionality reduction model. The reduced dimension $R$ is significantly smaller than $N$ and typically comparable to $\hat{M}_k$. For this purpose, singular value decomposition (SVD) is utilized to decompose the non-square matrix $\boldsymbol{\hat{Y}}_k$ into three matrices $\boldsymbol{A} \in\mathbb{R}^{N \times N} $, $\boldsymbol{S} \in \mathbb{R}^{N \times \hat{M}_k}$, and $\boldsymbol{B} \in\mathbb{R}^{\hat{M}_k \times \hat{M}_k} $, as follows.
\begin{equation}
\boldsymbol{\hat{Y}}_k = \boldsymbol{A}
\boldsymbol{S}  \boldsymbol{B}^{\text{T}} =
\left[\begin{array}{ccccc}
\alpha_{11}  &   \ldots & \alpha_{1R} &\ldots &\alpha_{1N} \\
\vdots &  \cdots  &\vdots  &\cdots  & \vdots \\
\alpha_{N1} &   \ldots & \alpha_{NR} &\ldots & \alpha_{NN}
\end{array}\right] 
\left[\begin{array}{cccc}
\sqrt{\lambda_1} & 0 & \ldots & 0 \\
0& \sqrt{\lambda_2} & \ldots & 0 \\
\vdots & \vdots & \ddots & \vdots \\
0 & 0 & \ldots & \sqrt{\lambda_{\hat{M}_k}} \\
\vdots & \vdots &  & \vdots \\
0 & 0 & \ldots & 0 
\end{array}\right] 
\left[\begin{array}{ccc}
\beta_{11}  &   \ldots & \beta_{1\hat{M}_k} \\
\vdots &  \cdots  & \vdots \\
\beta_{\hat{M}_k 1} &   \ldots & \beta_{\hat{M}_k \hat{M}_k}
\end{array}\right] 
\label{eq_svd}
\end{equation}
where $\boldsymbol{A}$ and $\boldsymbol{B}$ are orthogonal matrices, $\boldsymbol{S}$ denotes the singular value matrix with only diagonal elements $\sqrt{\lambda_i}$ being non-zero and arranged in descending order, and $\lambda_i$ represent non-zero eigenvalues of $\boldsymbol{\hat{Y}}_k  \boldsymbol{\hat{Y}}_k^{\text{T}}$. By using a truncated matrix $\boldsymbol \Lambda$ formed by extracting the first $R$ columns from the left eigenvector matrix $\boldsymbol{A}$, the low-rank approximation for $\boldsymbol{\hat{Y}}_k$ is achieved \citep{horn2012matrix} as: 
\begin{equation}
\boldsymbol{\hat{Y}}_k^\downarrow = \mathcal{R}^\downarrow \left( \boldsymbol{\hat{Y}}_k \right) = \boldsymbol \Lambda ^{\text T } \boldsymbol{\hat{Y}}_k,
\quad
\boldsymbol \Lambda  =\left[\begin{array}{ccc}
\alpha_{11}  &   \ldots & \alpha_{1R} \\
\vdots &  \cdots    & \vdots \\
\alpha_{N1} &   \ldots & \alpha_{NR} 
\end{array}\right] 
\label{Dim-Reduction}
\end{equation}
where $\boldsymbol{\hat{Y}}_k^\downarrow$ is the projection of the original physical field $\boldsymbol{\hat{Y}}_k$ onto the subspace $\boldsymbol \Lambda$ generated by $R$ principal axes. Dimensionality reduction is carried out individually on each cluster, and a reduced-order dataset, $\mathbb D (\boldsymbol{X} , \boldsymbol{Y}^\downarrow )$, is constructed from $ \mathbb D ( \boldsymbol{X} , \boldsymbol{Y} )$. The $K$ clustered reduced-order models read:
\begin{equation}
\mathbb D (\boldsymbol{X} , \boldsymbol{Y}^\downarrow ) = \mathbb D ( \boldsymbol{\hat{X}}_1 , \boldsymbol{\hat{Y}}_1^\downarrow ) \cup \mathbb D( \boldsymbol{\hat{X}}_2 , \boldsymbol{\hat{Y}}_2 ^\downarrow ) \cup \ldots \cup \mathbb D (\boldsymbol{\hat{X}}_K , \boldsymbol{\hat{Y}}_K^\downarrow )
\label{eq_xyreduced}
\end{equation}

\subsubsection{Construction of the Gaussian process sub-regressor}
\label{subsecGP}

The multi-pattern Gaussian process regressor consists of $K$ sub-regressors corresponding to $K$ clusters.
For the $k$-th cluster, the sub-regressor utilizes the collected input and reduced-order output data as follows:
\begin{equation}
\boldsymbol{\hat{X}}_k  =
\left[\begin{array}{c}
\hat{\chi}_1  \\
\vdots  \\
\hat{\chi}_{\hat{M}_k}
\end{array}\right],\quad
\boldsymbol{\hat{Y}}^\downarrow_k  =
\left[
\begin{array}{cccc}
\hat{\varphi}_{11}  & \hat{\varphi}_{12}  &  \ldots & \hat{\varphi}_{1\hat{M}_k} \\
\vdots &  \vdots  &\cdots  &\vdots  \\
\hat{\varphi}_{R1} &  \hat{\varphi}_{R2}   &  \ldots & \hat{\varphi}_{R\hat{M}_k}
\end{array}
\right] 
\end{equation}
where $\hat{\varphi}$ denotes latent variables from the dimensionality reduction. The $r$-th output latent variable within the reduced-order model, namely the transpose of the $r$-th row in $\boldsymbol{\hat{Y}}_k^\downarrow$, follows a prior joint Gaussian distribution with zero mean as follows. 
\begin{equation}
\boldsymbol{\hat{\varphi}}_r =
\left[\begin{array}{c}
\hat{\varphi}_{r1}\\
\vdots  \\
\hat{\varphi}_{r\hat{M}_k}
\end{array}\right]
 \sim \mathcal{N} \left( \boldsymbol{0} , \boldsymbol{K}(\boldsymbol{\hat{X}}_k,\boldsymbol{\hat{X}}_k) +\sigma_\text n ^2 \boldsymbol I \right) 
\label{eq_gp}
\end{equation}
where $\sigma_\text n$ denotes the inherent noise from training data, which is assumed following a Gaussian distribution, and $\boldsymbol{I}$ denotes an identity matrix. The $\boldsymbol K$ represents the covariance matrix and is given as:
\begin{equation}
\boldsymbol{K}(\boldsymbol{\hat{X}}_k,\boldsymbol{\hat{X}}_k)=
\left[\begin{array}{ccc}
\kappa \left(\hat{\chi}_1,\hat{\chi}_1\right)  &   \ldots & \kappa \left(\hat{\chi}_1,\hat{\chi}_{\hat{M}_k} \right)  \\
\vdots &  \cdots  & \vdots \\
\kappa \left(\hat{\chi}_{\hat{M}_k},\hat{\chi}_1 \right) &   \ldots & \kappa \left(\hat{\chi}_{\hat{M}_k},\hat{\chi}_{\hat{M}_k} \right) 
\end{array}\right] 
    \label{eq_K}
\end{equation}
where $\kappa$ is the kernel function, quantifying the covariance among outputs. 
It is formulated as a function of inputs. 

In this study, the squared exponential (SE) kernel, also known as the radial basis function (RBF) or Gaussian kernel, is employed as follows. 
\begin{equation}
    \kappa \left( \hat{\chi}_i, \hat{\chi}_j \right) = \sigma_{\text f}^2 \exp \left( -\frac{ |\hat{\chi}_i - \hat{\chi}_j |^2}{2\ell^2} \right) 
    \label{eq_kappa}
\end{equation}
where $\sigma_\text f$ represents the signal variance and $\ell$ denotes the characteristic length. It is clear that $\kappa \left( \hat{\chi}_i, \hat{\chi}_j \right)$ approaches unity for variables with closely related inputs and diminishes with increasingly separated inputs. 

As shown in Eqs.~\autoref{eq_gp}, \autoref{eq_K} and \autoref{eq_kappa} for the Gaussian process, the unknown parameters are $\sigma_\text n$ ,$\sigma_\text f$, and $\ell$. They are the so-called hyperparameters of the GPR model.
The characteristic length $\ell$ quantifies the distance in the parameter space at which outputs become less correlated, i.e., the larger the value of $\ell$, the smoother transitions in $\boldsymbol{\hat{\varphi}}_r$ within $\boldsymbol{\hat{X}}_k$. 
The signal variance $\sigma_\text f$ governs the range of the outputs, and $\sigma_\text n$ denotes the noise level and can be regarded as regularization to avoid overfitting.
Therefore, these hyperparameters have significant affects on the model's online performance and should be determined in a rigorous way. 
In this study, the maximizing marginal likelihood strategy is employed to fine-tune these parameters as follows. 

Given the $r$-th latent variable of the $k$-th cluster, the marginal likelihood, i.e., the conditional probability density of $\boldsymbol{\hat{\varphi}}_r$, is formulated as:
\begin{equation}
p\left( \boldsymbol{\hat{\varphi}}_r  \mid \boldsymbol{\hat{X}}_k, \boldsymbol{\theta}\right)=
\frac{1}{ \sqrt{ (2 \pi) ^ {\hat{M}_k} | \boldsymbol G } | }  
\exp \left( -\frac{1}{2} \boldsymbol{\hat{\varphi}}_r^\text T  \boldsymbol G ^{-1} \boldsymbol{\hat{\varphi}}_r \right)
\end{equation}
where $\boldsymbol \theta = \left[\ell ,\sigma_\text f, \sigma_\text n  \right]^\text T $, and $\boldsymbol G$ is defined as:
\begin{equation}
\boldsymbol G  = \boldsymbol{K}(\boldsymbol{\hat{X}}_k,\boldsymbol{\hat{X}}_k) +\sigma_\text n ^2 \boldsymbol I
\label{eq_G}
\end{equation}
Then, the negative log marginal likelihood function reads:
\begin{equation}
\mathcal{Q}(\boldsymbol{\theta}) = - \log
p\left( \boldsymbol{\hat{\varphi}}_r  \mid \boldsymbol{\hat{X}}_k, \boldsymbol{\theta}\right)=
\frac{1}{2} \boldsymbol{\hat{\varphi}}_r^\text T  \boldsymbol G ^{-1} \boldsymbol{\hat{\varphi}}_r 
+\frac{1}{2} \log |   \boldsymbol G  |
+\frac{\hat{M}_k}{2} \log 2\pi
\end{equation}
where the first term on the right-hand-side, containing training data, describes the empirical risk, and the second term associated with the covariance structure increases with the model complexity, signifying the model risk. 
By minimizing the objective function $\mathcal Q$, the process of maximizing marginal likelihood is thus transformed into an unconstrained optimization of variables $\boldsymbol \theta$ as follows.
\begin{equation}
\boldsymbol{\theta} =\underset{\boldsymbol{\theta}}{\arg \min } \; \mathcal{Q} (\boldsymbol{\theta}) 
\label{eq_theta}
\end{equation}

For optimizing the hyperparameters given by Eq.~\autoref{eq_theta}, the conjugate gradient descent method \citep{shewchuk1994introduction} is utilized. An initial search direction is determined as:
\begin{equation}
\boldsymbol{g}^{0} = -\nabla \mathcal Q  ^{0} 
\label{eq_direc}
\end{equation}
where $\nabla \mathcal Q$ represents the gradient of the objective function with respect to hyperparameters, and the superscripts denote iteration steps. In the $m$-th iteration, hyperparameters are updated as:
\begin{equation}
\boldsymbol{\theta}^{m+1} = \boldsymbol{\theta}^{m} + \tau^{m} \boldsymbol{g}^{m} 
\label{eq_thetaup}
\end{equation}
with $\tau^{m}$ being the step length, determined by a line search. The searching direction, $\boldsymbol{g}$, is updated according to the Polak-Ribière formula:
\begin{equation}
\boldsymbol{g}^{m+1} = -\nabla  \mathcal Q  ^{m+1}  + \frac{ ( \nabla  \mathcal Q^{m+1} )^{\text T} (\nabla  \mathcal Q^{m+1} - \nabla \mathcal Q^{m} ) } {  ( \nabla  \mathcal Q^{m})^{\text T} \nabla \mathcal Q^{m}  }\boldsymbol{g}^{m}
\label{eq_gmp1}
\end{equation}
The partial derivatives within $\nabla \mathcal Q$ are detailed as follows: 
\begin{equation} \left\{
\begin{aligned} 
& \frac{\partial \mathcal Q}{\partial \ell }= \frac{1}{2} \boldsymbol{\hat{\varphi}}_r^{\text{T}} \boldsymbol{G}^{-1} \frac{\partial \boldsymbol{G}}{\partial \ell} \boldsymbol{G}^{-1} \boldsymbol{\hat{\varphi}}_r +\frac{1}{2} \operatorname{tr}\left(\boldsymbol{G}^{-1} \frac{\partial \boldsymbol{G}}{\partial \ell}\right) \\
& \frac{\partial \mathcal Q}{\partial \sigma_\text f } = \frac{1}{2} \boldsymbol{\hat{\varphi}}_r^{\mathrm{T}} \boldsymbol{G}^{-1} \frac{\partial \boldsymbol{G}}{\partial \sigma_\text f} \boldsymbol{G}^{-1} \boldsymbol{\hat{\varphi}}_r+\frac{1}{2} \operatorname{tr}\left(\boldsymbol{G}^{-1} \frac{\partial \boldsymbol{G}}{\partial \sigma_\text f}\right) \\
& \frac{\partial \mathcal Q}{\partial \sigma_\text n } = \frac{1}{2} \boldsymbol{\hat{\varphi}}_r^{\mathrm{T}} \boldsymbol{G}^{-1} \frac{\partial \boldsymbol{G}}{\partial \sigma_\text n} \boldsymbol{G}^{-1} \boldsymbol{\hat{\varphi}}_r+\frac{1}{2} \operatorname{tr}\left(\boldsymbol{G}^{-1} \frac{\partial \boldsymbol{G}}{\partial \sigma_\text n}\right)
\end{aligned} \right.
\label{eq_gradQ}
\end{equation}
The hyperparameters are optimized by iterating through Eqs.~\autoref{eq_thetaup} and \autoref{eq_gmp1} until convergence. 

It is important to point out that, for each cluster, all $R$ latent variables share the same set of optimized hyperparameters. However, separate optimization is conducted for each cluster, ensuring tailored performance for distinct patterns.

\subsection{Online prediction}
\label{subonline}

For unseen inputs, the online prediction of the well-trained multi-pattern Gaussian process regressor includes three steps, adhering to the divide-and-conquer strategy. They are the pre-prediction classification, prediction, and dimensionality restoration in sequence.   

\subsubsection{Pre-prediction classification}

Before prediction, the k-nearest neighbors (KNN) method is employed to categorize the unseen inputs into previously defined clusters. 
For the training dataset presented in Eq.~\autoref{eq_finalXY} with the sample size of $M$, the input parameters read: 
\begin{equation}
\boldsymbol X = \left[ \chi_{1} , \; \chi_{2} , \; \cdots , \;\chi_{M} \right]^{\text T}
\end{equation}
The cluster indices for these samples are denoted by the following vector.
\begin{equation}
\boldsymbol \Gamma = \left[ \gamma_{1} , \; \gamma_{2} , \; \cdots , \;\gamma_{M} \right]^{\text T},  \quad \gamma_i \in [1,K]
\end{equation}
with $\gamma_i$ indicating the index of cluster, to which the $i$-th sample belongs. 

For an unseen instance $\chi^*$, its proximity to each $\chi_i$ in $\boldsymbol{X}$ is computed using the Euclidean distance, yielding a neighborhood set $\mathbb K$ of $K_\text {knn}$ nearest neighbors. The corresponding cluster index $\gamma^*$, is determined based on the majority voting among these neighbors. It is expressed as the following equation:
\begin{equation}
\gamma^* =\underset{j }{\arg \max } \; \textstyle \sum _{\chi_i \in \mathbb K } \delta (\gamma_i = j) , \quad i \in [1, M], \quad j \in [1, K] 
\label{eq_knn}
\end{equation}
where $\delta$ acts as a binary function indicating agreement between $\gamma_i$ and $j$. 

Likewise, one can determine the cluster IDs for a set of $M^*$ cases by using Eq.~\autoref{eq_knn}. 
For the set of  $M^*$ cases, the input parameters are represented as:
\begin{equation}
\boldsymbol X^* = \left[ \chi_{1}^* , \; \chi_{2}^* , \; \cdots , \;\chi_{M^*}^* \right]^{\text T} 
\end{equation}
Correspondingly, the cluster indicator vector is obtained as:
\begin{equation}
\boldsymbol \Gamma^* = \left[ \gamma_{1}^* , \; \gamma_{2}^* , \; \cdots , \;\gamma_{M^*}^* \right]^{\text T},  \quad \gamma_i^* \in [1,K]
\end{equation}
Finally, the input data $\boldsymbol X^* $ is then partitioned into $K^*$ non-overlapping subsets, where the $K^*$ is less than or equal to the total cluster number $K$, depending on the distribution of input parameters.
Consequently, $\boldsymbol X^* $ can be rewritten as:
\begin{equation}
 \boldsymbol{X}^*   =   \boldsymbol{\hat{X}}_1^*  \cup  \boldsymbol{\hat{X}}_2^*   \cup \ldots \cup  \boldsymbol{\hat{X}}_{K^*}^* 
 \label{eq_xstar}
\end{equation} 

\subsubsection{Prediction with multi-pattern Gaussian process regressor}

The instances in Eq.~\autoref{eq_xstar} are predicted independently for each cluster.
Specifically, for the input $\boldsymbol{\hat{X}}_k^*$ belonging to the $k$-th cluster, the output $\boldsymbol{\hat{\varphi}}_r^\ast$ and the known $\boldsymbol{\hat{\varphi}}_r$ for the $r$-th latent variable from the training set follow a prior joint Gaussian distribution, as follows.
\begin{equation}
\left[\begin{array}{c}
\boldsymbol{\hat{\varphi}}_r \\
\boldsymbol{\hat{\varphi}}_r^\ast
\end{array}\right] \sim \mathcal{N}\left(\boldsymbol{0},\left[\begin{array}{cc}
\boldsymbol G & \boldsymbol K\left(\boldsymbol{\hat{X}}_k, \boldsymbol{\hat{X}}_k^*\right) \\
\boldsymbol K\left(\boldsymbol{\hat{X}}_k^*, \boldsymbol{\hat{X}}_k\right) & \boldsymbol K\left(\boldsymbol{\hat{X}}_k^*, \boldsymbol{\hat{X}}_k^*\right)
\end{array}\right]\right)
\end{equation}
The posterior distribution of the estimated parameters $\boldsymbol{\hat{\varphi}}_r^\ast$ is obtained after employing Eqs.~\autoref{eq_gp}, \autoref{eq_K}, \autoref{eq_kappa}, and \autoref{eq_G}, alongside the hyperparameter determination from Eq.~\autoref{eq_theta}:
\begin{equation}
\boldsymbol{\hat{\varphi}}_r^\ast \mid \boldsymbol{\hat{X}}_k^*, \boldsymbol{\hat{X}}_k, \boldsymbol{\hat{\varphi}}_r 
\sim
\mathcal N 
\left(   \boldsymbol K\left(\boldsymbol{\hat{X}}_k^*, \boldsymbol{\hat{X}}_k\right)  \boldsymbol{G}^{-1} \boldsymbol{\hat{\varphi}}_r, \;
\boldsymbol K\left(\boldsymbol{\hat{X}}_k^*, \boldsymbol{\hat{X}}_k^*\right)
-\boldsymbol K\left(\boldsymbol{\hat{X}}_k^*, \boldsymbol{\hat{X}}_k\right)
\boldsymbol{G}^{-1} \boldsymbol K\left(\boldsymbol{\hat{X}}_k, \boldsymbol{\hat{X}}_k^*\right) \right)
\label{eq_gprpre}
\end{equation}
where the predicted mean value is 
\begin{equation}
\boldsymbol{\hat{\bar{\varphi}}}^\ast_r  =  \boldsymbol K\left(\boldsymbol{\hat{X}}_k^*, \boldsymbol{\hat{X}}_k\right)  \boldsymbol{G}^{-1} \boldsymbol{\hat{\varphi}}_r 
\label{eq_predict}
\end{equation}
and the predicted variance is 
\begin{equation}
\boldsymbol \hat{\sigma}^*_{kr}   = \boldsymbol K\left(\boldsymbol{\hat{X}}_k^*, \boldsymbol{\hat{X}}_k^*\right)
-\boldsymbol K\left(\boldsymbol{\hat{X}}_k^*, \boldsymbol{\hat{X}}_k\right)
\boldsymbol{G}^{-1} \boldsymbol K\left(\boldsymbol{\hat{X}}_k, \boldsymbol{\hat{X}}_k^*\right) 
\label{eq_sigma}
\end{equation}
Eq.~\autoref{eq_predict} serves for making predictions, while Eq.~\autoref{eq_sigma} yields the associated confidence interval. 
By recursively applying Eq.~\autoref{eq_predict} for $r$ ranging from 1 to $R$ with consistent hyperparameters, a reduced-order result can be obtained as follows.
\begin{equation}
\boldsymbol{\hat{\bar{Y}}}^{\ast\downarrow}_k =
\left[
\boldsymbol{\hat{\bar{\varphi}}}_1 ^{\ast } ,\;
\ldots ,\;
\boldsymbol{\hat{\bar{\varphi}}}_R ^{\ast  } 
\right] ^\text T
\end{equation}
The matrix $\boldsymbol{\hat{\bar{Y}}}^{\ast\downarrow}_k$ contains elements that correspond to the predicted mean values of latent variables, residing in the real number space $\mathbb R ^{R \times \hat{M}^*_k}$ with $\hat{M}^*_k$ being the number of cases to be predicted in the $k$-th cluster. 

\subsubsection{Physical fields reconstruction}

In the final step, the meaningful physical fields that represent the structural behaviors are reconstructed from the predicted reduced-order variables $\boldsymbol{\hat{\bar{Y}}}^{\ast\downarrow}_k$. 
To this end, the reverse implementation as shown in Eq.~\autoref{Dim-Reduction} is conducted on $\boldsymbol{\hat{\bar{Y}}}^{\ast\downarrow}_k$ as follows.
\begin{equation}
\boldsymbol{\hat{\bar{Y}}}^{\ast}_k   = \mathcal{R}^\uparrow \left( \boldsymbol{\hat{\bar{Y}}}^{\ast\downarrow}_k \right) = \boldsymbol \Lambda \boldsymbol{\hat{\bar{Y}}}^{\ast\downarrow}_k
\label{eq_restoration}
\end{equation}
where $\mathcal{R}^\uparrow$ represents the dimensionality restoration, using the matrix $\boldsymbol \Lambda$ obtained in Eq.~\autoref{Dim-Reduction}. The notation $\boldsymbol{\hat{\bar{Y}}}^{\ast}_k$ represents the physical field within the $k$-th cluster, which has been expanded back to an $\mathbb R ^{N \times \hat{M}^*_k}$ space. 

It is important to point out that each scalar physical field (such as damage) goes through an individual compression, training, prediction, and restoration processes. Vector fields (like displacement) and tensor fields (such as stress and strain) are also treated separately, considering each component as an individual scalar field.

\subsection{Overall procedure}

\begin{figure}[tbph]
    \begin{centering}
    \includegraphics[width=13cm]{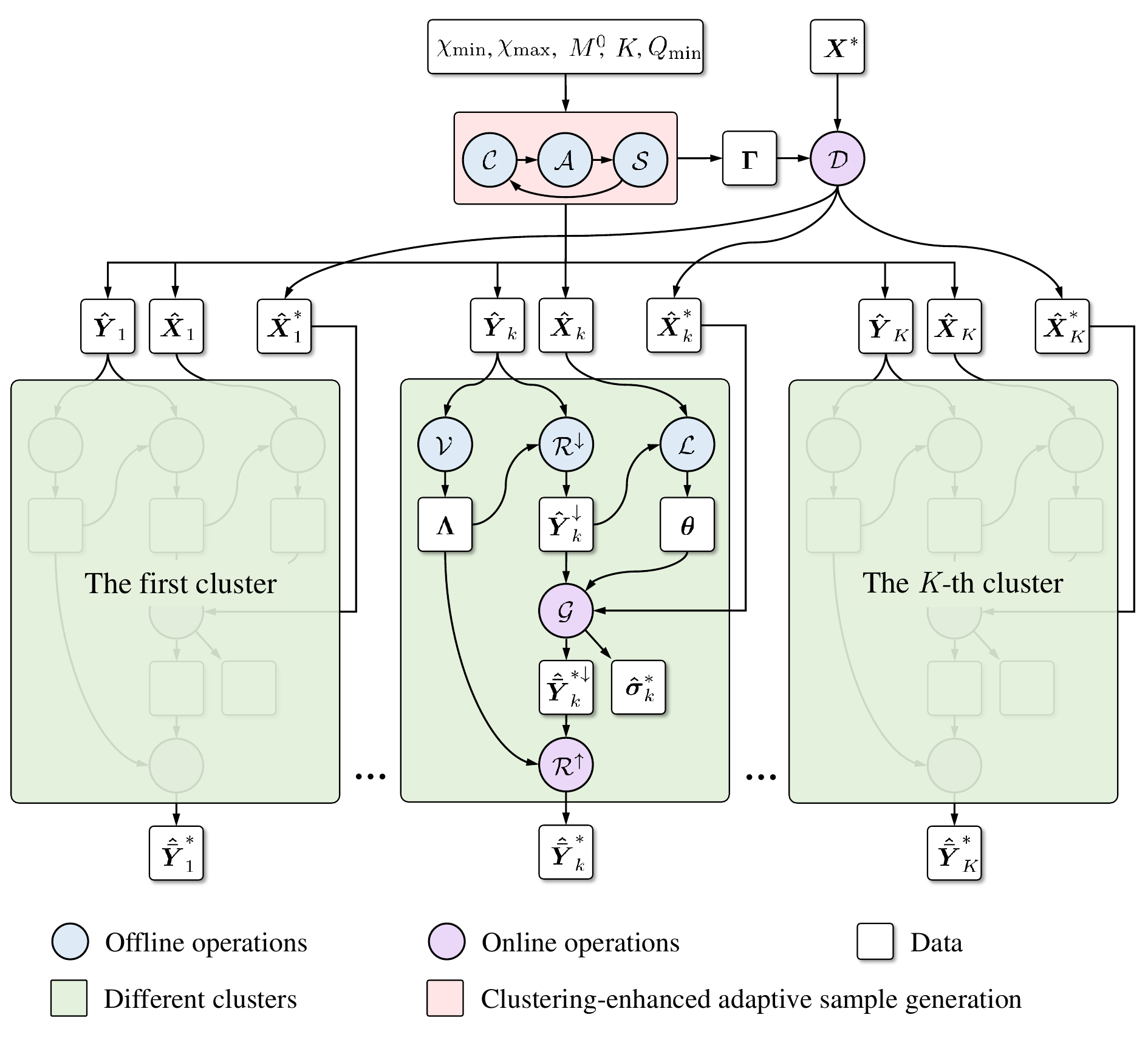}
    \par\end{centering}
    \caption{Implementation procedure of the proposed CAG method, where the operator $\mathcal C$ refers to the clustering process detailed in Eqs.~\autoref{eq_initcent}, \autoref{eq_edist}, \autoref{eq_put} and \autoref{eq_updatecent}, the operator $\mathcal A$ for the adaptive sample addition according to Eq.~\autoref{eq_xiadd}, the operator $\mathcal S$ for the high-fidelity simulation solver, the operator $\mathcal L$ for the optimization process, the operator $\mathcal V$ for Eq.~\autoref{eq_svd}, the operator $\mathcal D$ for pre-prediction classification in Eq.~\autoref{eq_knn} and the operator $\mathcal G$ for the online predictions.}
\label{fig_pro}
\end{figure}

As depicted in \autoref{fig_pro}, the overall implementation procedure of the proposed CAG method consists of six steps as detailed below.

\begin{enumerate}[(1)]
    \item Generate $K$ clustered training datasets that cover the parameter space with  the clustering-enhanced adaptive sample generation scheme. 

    \item Construct reduced-order datasets for each cluster through the dimensionality reduction technique denoted by $\mathcal{R}^\downarrow$, utilizing the SVD method given by Eq.~\autoref{eq_reduce}.

    \item Train the multi-pattern Gaussian process regressor independently on each cluster, with optimizd hyperparameters using Eqs.~\autoref{eq_theta}, \autoref{eq_direc}, \autoref{eq_thetaup}, \autoref{eq_gmp1} and \autoref{eq_gradQ}. 

    \item For unseen inputs $\boldsymbol{X}^*$, utilize pre-prediction classification $\mathcal D$ expressed by Eq.~\autoref{eq_knn} to assign prediction tasks into appropriate clusters.

    \item Conduct online predictions for each cluster to obtain predicted reduced-order latent variables according to Eqs.~\autoref{eq_predict} and \autoref{eq_sigma}.

    \item Finally, reconstruct physical fields from reduced-order models using dimensionality restoration, i.e., Eq.~\autoref{eq_restoration}.
\end{enumerate}

\section{Numerical examples}
\label{secCase}

To demonstrate the accuracy and efficiency of the proposed method for predicting problems with multiple patterns, we have studied six representative problems with diverse patterns. These problems involve a range of input parameters and cover both static and dynamic problems with material, geometric, and boundary condition (contact) nonlinearities. 

\subsection{Regression on a nonlinear function with highly localized features}main

A mathematically complex function with pronounced nonlinearity and highly localized features is addressed using the CAG method to showcase its advantages over the traditional GPR with an equal number of samples. 
It is a wavelet-like function defined as:
\begin{equation}
w(t) = 1+ \sin(6t) \cdot e^{ -\frac{1}{2}t^2 }
\label{eq_wavelet}
\end{equation}
This function borrows the concept from wavelet functions, characterized by prominent, non-uniform amplitude oscillations occurring within a localized domain. The constant of one to the right hand side of Eq.~\autoref{eq_wavelet} is applied to maintain a non-zero equilibrium point and to avoid singularities during the relative error calculation. 

For this example, $t$ and $w$ are taken as input-output pairs. 
The input range is set as $[-15, 15]$.
The proposed method takes the settings of $K=3$, $Q_\text{min}=5$, and the initial sample size $M^0$ spans from 5 to 250, resulting in a final $M$ ranging from 23 to 250. For large initial values of $M^0$, the convergence condition in Eq.~\autoref{eq_Msample} is met during the first clustering step without invoking $\mathcal{A}$, resulting the same $M$ with $M^0$. 
Eq.~\autoref{eq_wavelet} is directly used as the solver $\mathcal{S}$ for dataset creation. 
It should be noted that the dimensionality reduction or restoration is not necessary in this case as $N$ equals 1. 
The prediction includes $M^* = 1,000$ cases within the range of $[-15, 15]$. 
In addition, a conventional GPR model is employed for regression of the same problem, utilizing 20 to 250 uniformly distributed training samples from the same range for comparison. The prediction accuracy is evaluated by calculating the maximum relative error between the predicted $w$ values and those obtained analytically from Eq.~\autoref{eq_wavelet}. 

\begin{figure}[tbph]
    \begin{centering}
    \includegraphics[width=14.5cm]{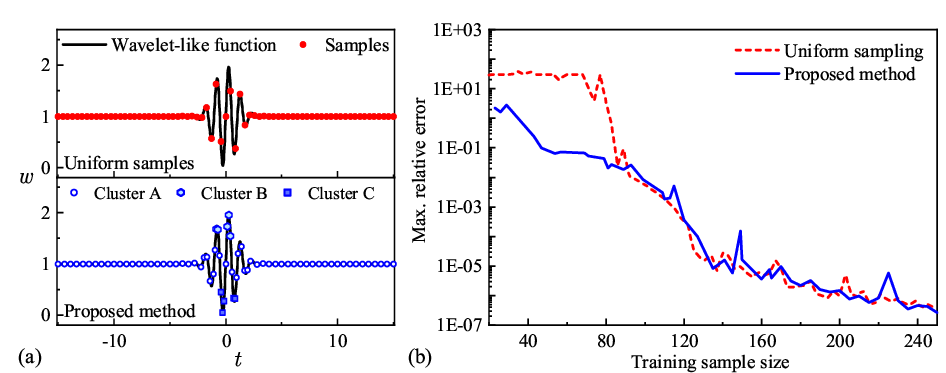}
    \par\end{centering}
    \caption{Comparison between the proposed CAG method and the GPR with uniform sampling: (a) The sample distribution with 71 training samples; (b) Variation of maximum relative error with different training sample sizes.}
\label{fig_wavelet_error}
\end{figure}

\autoref{fig_wavelet_error}(a) compares the distributions of 71 samples for the proposed method with the conventional GPR. The comparison demonstrates that the proposed method's training samples accurately capture the wavelet-like function's localized oscillating zone. In contrast, the traditional uniform sampling method fails to capture the maximum oscillation amplitude with the same sample size.
\autoref{fig_wavelet_error}(b) compares prediction performance across a broader scope, encompassing sample sizes ranging from 20 to 250. It is clear that the proposed method significantly outperforms the uniform sampling approach if the sample size is less than 80, achieving maximum relative error reductions of one to two orders of magnitude. As the sample size exceeds 100, both methods exhibit comparable and satisfactory levels of prediction accuracy.

\begin{figure}[tbph]
    \begin{centering}
    \includegraphics[width=16cm]{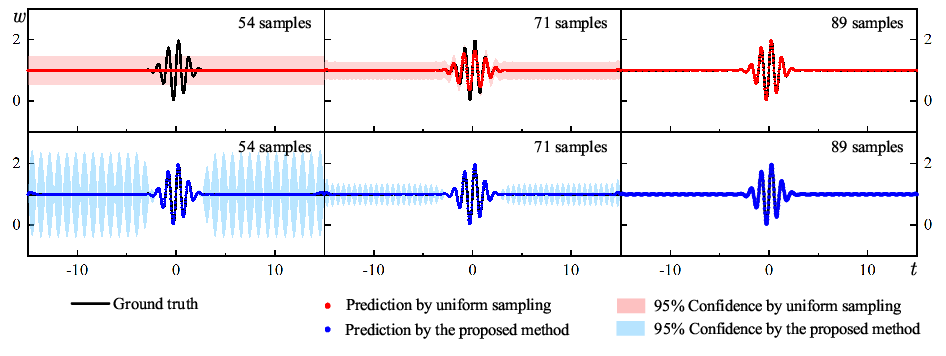}
    \par\end{centering}
    \caption{Regression of the wavelet-like function by the proposed CAG method and GPR using uniform sampling.}
\label{fig_wavelet_comp}
\end{figure}

\autoref{fig_wavelet_comp} demonstrates the regression of the wavelet-like function using both methods with 54, 71, and 89 training samples, where the 95\% confidence zone is depicted as well.
The comparison of the predicted results further emphasizes the proposed method's proficiency in capturing the function's localized characteristics with limited samples. It underscores the proposed method's ability to adapt the sample distribution to the input parameter space, thereby enhancing prediction accuracy, particularly for potential scenarios exhibiting highly localized features, which is common in solid mechanics problems, as shown in the following sections.

\subsection{Vibration of a mass-spring-damper system}

\begin{figure}[tbph]
    \begin{centering}
    \includegraphics[width=5.5cm]{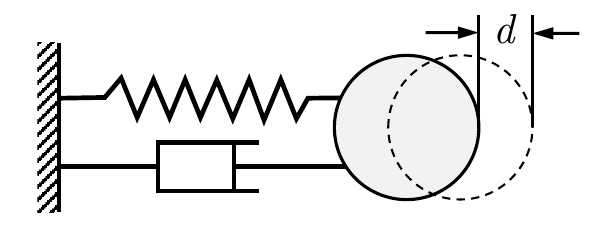}
    \par\end{centering}
    \caption{Schematic diagram for free vibration of the mass-spring-damper system.}
\label{fig_spring_sk}
\end{figure}

The second example focuses on predicting the time series of a mass-spring-damper system experiencing free vibration, as shown in \autoref{fig_spring_sk}. Drawing upon prior knowledge, we understand that vibrations can manifest in various patterns, such as underdamped and overdamped vibrations. This numerical example aims to demonstrate the proposed method's capability to depict varying patterns over time.

The governing equation for the vibration reads:
\begin{equation}
m_\text s \frac{\partial ^2 d }{\partial t ^2} + c_\text s \frac{\partial d }{\partial t } + s_\text s d = 0
\label{eq_spring_govern}
\end{equation}
where $m_\text s$ is the mass, $s_\text s$ is the spring stiffness, $c_\text s$ is the damping coefficient, and $d$ is the displacement and takes the value of 0 for the equilibrium position. By applying the initial condition
\begin{equation}
\left. d \right|_{t=0} = d_0, \quad \left. \frac{\partial d }{\partial t } \right|_{t=0} = 0
\end{equation}
with $d_0$ being the initial displacement, the analytical solution to Eq.~\autoref{eq_spring_govern} is given as: 
\begin{equation}
d=\left\{\begin{array}{lc}
{e}^{-\zeta \omega_\text n t}\left(d_0 \cos \omega_\text d t+\dfrac{\zeta \omega_\text n d_0}{\omega_d} \sin \omega_\text d t\right) & \zeta<1 \\
{e}^{-\omega_\text n t}\left(d_0+d_0 \omega_n t\right) & \zeta=1 \\
\dfrac{-\left|\varpi_2 \right| }{\left|\varpi_1\right|-\left|\varpi_2\right|} d_0 {e}^{-\left|\varpi_1\right| t}+\dfrac{\left|\varpi_1\right| }{\left|\varpi_1\right|-\left|\varpi_2\right|} d_0 {e}^{-\left|\varpi_2\right| t} & \zeta>1
\end{array}\right.
\label{eq_spring_sol}
\end{equation}
where $\zeta = c_\text s/(2\sqrt{s_\text s m_\text s})$ is the damping ratio, $\omega_\text n = \sqrt{s_\text s/m_\text s}$ is the natural frequency, and $\omega_\text d$ and $\varpi$ are given, respectively, as:
\begin{equation}
\omega_\text d = \omega_\text n \sqrt{1-\zeta^2} , \quad 
\varpi_{1,2} = -(\zeta \mp \sqrt{\zeta^2 - 1}) \omega_\text n 
\end{equation}

In this case, the damping ratio $\zeta$ and displacement $d$ are designated as the input-output pairs. Initially, adaptive training samples are generated with parameters $K=3$, $Q_\text{min}=4$, and an input range of $[0, 2]$. Starting with an initial sample size of $M^0=5$, the final sample count reaches $M=16$. Eq.~\autoref{eq_spring_sol} serves as the solver $\mathcal{S}$, utilizing a mass of $m_\text s = 0.1 \;\rm Kg$, spring stiffness of $s_\text s=200 \;\rm N/m$, and initial displacement $d_0=0.1 \;\rm m$. Vibrations lasting 1 second are considered, and $N=200$ points are selected along the time axis. The reduced dimension is set to $R=50$. The prediction includes $M^* = 50$ cases within the same range. In addition, the prediction accuracy is compared with those obtained using traditional GPR with uniform sampling.

\begin{figure}[tbph]
    \begin{centering}
    \includegraphics[width=14.5cm]{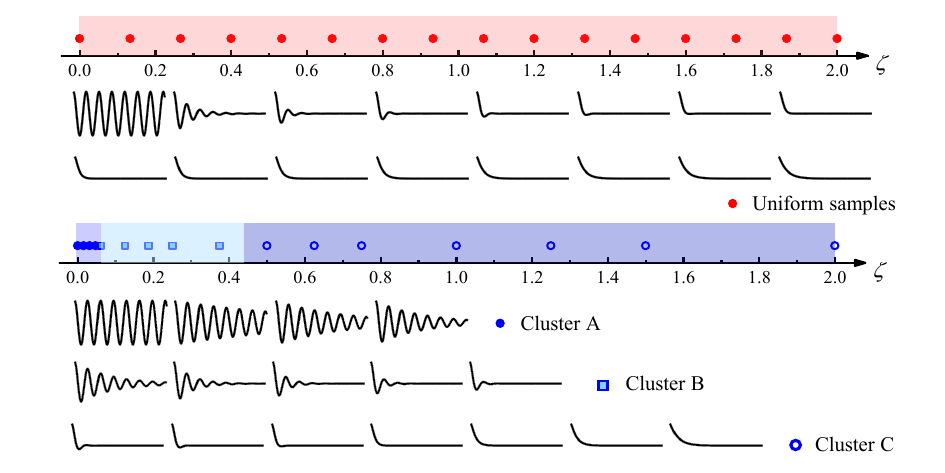}
    \par\end{centering}
    \caption{Comparison of the proposed CAG method and the GPR with uniform sampling for sample distribution and vibration patterns using 16 samples.}
\label{fig_sample_distribute}
\end{figure}

\autoref{fig_sample_distribute} displays the distributions of training samples along the $\zeta$ axis for both methods, along with their respective vibration patterns. It is evident that the uniform sampling method fails to capture the significant changes in vibration patterns observed for $\zeta < 0.2$, with only 2 of the 16 samples in this range, while others exhibit highly similar vibration patterns. In contrast, the proposed method deciphers the samples into three distinct clusters: Cluster A with four samples in the range of $[0,0.055)$, Cluster B with five samples in the range of $(0.055,0.438)$, and Cluster C with seven samples in the range of $(0.438,2]$. 
It suggests allocating more samples in regions where displacement response patterns undergo significant changes while maintaining sparsity for smoother transitions. Consequently, a more representative and appropriate distribution of samples is ensured.

\begin{figure}[tbph]
    \begin{centering}
    \includegraphics[width=16cm]{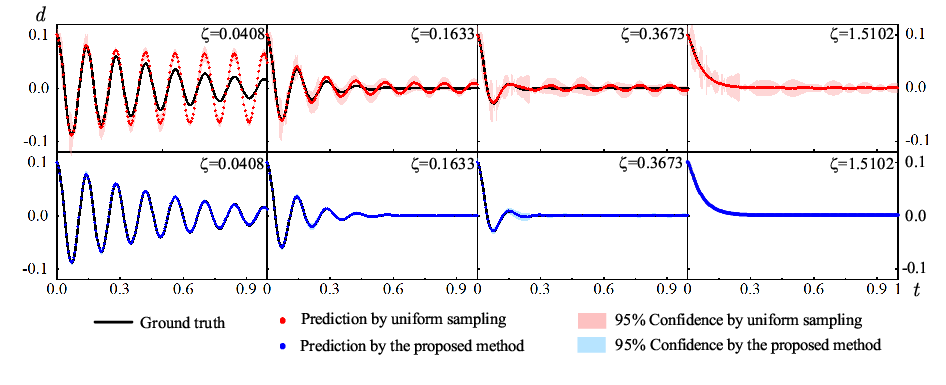}
    \par\end{centering}
    \caption{Comparison of the proposed CAG method and the GPR with uniform sampling for predicting vibration displacement and confidence intervals using 16 samples.}
\label{fig_spring_error}
\end{figure}

A comparative analysis of prediction results from both methods is presented in \autoref{fig_spring_error}, for four cases with the value of $\zeta$ equal to 0.0408, 0.1633, 0.3673, and 1.5102, respectively.
The proposed method demonstrates higher accuracy in displacement forecasts, along with tighter confidence intervals, indicating a higher degree of credibility in its predictions.
For the case of $\zeta=0.0408$, the uniform sampling method's prediction is influenced by the nearest training sample ($\zeta=0$), leading to an underestimation of amplitude decay. However, the proposed method offers precise predictions by allocating more samples to this region. 
Notably, even for sparse training samples such as the case of $\zeta=1.5102$, the proposed method maintains higher accuracy and narrower confidence intervals than the uniform sampling method. 
This is attributed to the pre-prediction classification process, ensuring that vibration patterns with larger amplitudes in Clusters A and B do not interfere with predictions in Cluster C.

Two supplementary cases with 23 and 63 samples are further conducted. The results are summarized in \autoref{tabpar}, offering a quantitative comparison of the maximum relative errors and mean square errors of displacement between the two methods.
It is demonstrated that the proposed method outperforms the GPR with the uniform sampling, with maximum relative errors reduced by 1 to 2 orders of magnitude and mean square errors lowered by 2 to 3 orders of magnitude for all cases.

\begin{table}
    \small
    \caption{Comparison of errors using the proposed CAG method and the GPR with uniform sampling in predicting vibration of the mass-spring-damper system.}
    \centering
    \begin{tabular}{ccccc}
        \toprule
        Sample size  & Method & Max. relative error &   Mean square error \\
        \midrule
        \multirow{2}{*}{16}  & Uniform sampling  &  61.70\%  & $2.38\times 10^{-5}$  \\
                             & Proposed method &   5.61 \%    &  $  1.20\times 10^{-7}$ \\
        \midrule
        \multirow{2}{*}{23}  & Uniform sampling  &  27.87\%  &   $4.08\times 10^{-6}$  \\
                             & Proposed method  &   0.89\%   &   $2.48\times 10^{-8}$\\
        \midrule
        \multirow{2}{*}{63}  & Uniform sampling  &  0.0336\%  &   $ 3.06 \times 10^{-12}$  \\
                             & Proposed method  &   0.00148\%   &   $9.00\times 10^{-15}$\\
        \bottomrule
    \end{tabular}
    \label{tabpar}
\end{table}

\subsection{Hyperelastic cylinder under axial stretching and compression}\label{CylinderProblem}

\begin{figure}[tbph]
    \begin{centering}
    \includegraphics[width=4.2cm]{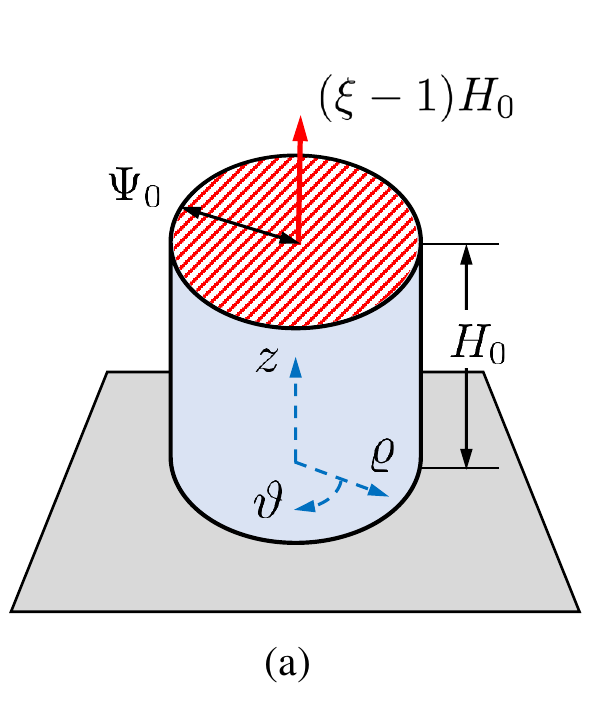} \quad
    \includegraphics[width=8.4cm]{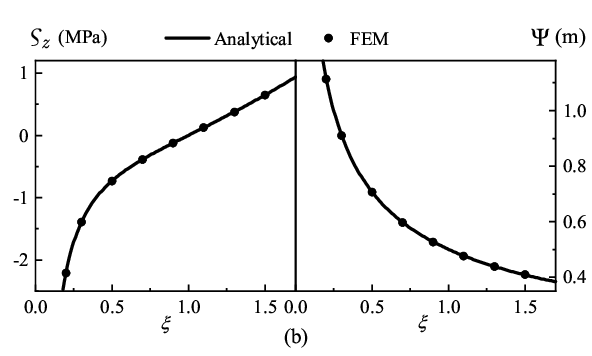}
    \par\end{centering}
    \caption{(a) Schematic diagram of the cylinder under the extension and compression loading; (b) Comparison of vertical stress and radius between simulation solver and the analytical solution.}
\label{fig_hyper_sch}
\end{figure}

The third example focuses on the deformation prediction of a hyperelastic cylinder under axial loading, taking into account both material and geometric nonlinearities, to demonstrate the accuracy and the generalization capability of the proposed method. 
 As illustrated in \autoref{fig_hyper_sch} (a), the cylinder has an initial radius of $\Psi_0 = 0.5 \; \rm m$ and a height of $H_0=1 \;\rm m$, and the axial stretching and compression loads are considered. 

 A cylindrical coordinate system $(\varrho,\vartheta,z)$ is used to describe this problem, and boundary conditions for vertical displacement $d_z$ are given by:
\begin{equation}
d_z|_{z = H_0} = (\xi-1)H_0, \quad d_z|_{z = 0} = 0
\end{equation}
where $\xi$ represents the axial stretch ratio, with $\xi<1$ denoting compression and $\xi>1$ indicating stretching. The material of the cylinder is incompressible, isotropic and hyperelastic. The two-parameter Mooney-Rivlin constitutive law \citep{mooney1940theory,rivlin1948large} is adopted for the material, and the strain energy function $W$ is given by:
\begin{equation}
W=\mathscr D _1\left(I_1-3\right)+ \mathscr D_2\left(I_2-3\right)
\end{equation}
where $I_1$ and $I_2$ are the first and second invariants of Cauchy-Green deformation tensor, respectively. The material parameters $\mathscr D_1$ and $\mathscr D_2$ take the values of 0.200912 MPa and 0.004235 MPa, respectively. 

In the absence of frictional forces acting on the cylinder's ends and allowing $d_\varrho$ to be unrestrained, the analytical solution to this problem is presented by Valiollahi et al. \citep{valiollahi_closed_2019}. In this case, the deformed radius $\Psi$ derived from the incompressible condition remains constant throughout the cylinder, and reads:
\begin{equation}
\Psi = \xi^{-1 / 2} \Psi _0
\end{equation}
The only non-zero stress component is the $z$-direction stress $\varsigma_z$, which remains constant throughout the cylinder and is expressed as:
\begin{equation}
\varsigma_{z}=2\left(\xi^2-\xi^{-1}\right) \mathscr{D}_1+2\left(\xi-\xi^{-2}\right) \mathscr{D}_2
\end{equation}
The analytical solution is utilized to validate the numerical simulation solver $\mathcal S$, e.g., the finite element method (FEM) software ANSYS, which is applied to construct the training dataset. The numerical results, obtained using a 1/4 model with 3,317 nodes, are plotted in \autoref{fig_hyper_sch}(b). By comparing stress and deformed radius across varying axial stretch ratios, it is demonstrated that the numerical results are in good agreement with the analytical solution. 

With the validated numerical simulation solver $\mathcal S$, a more complex mechanical behavior of the cylinder under the additional boundary conditions is considered to demonstrate the performance of the proposed method, as follows.
\begin{equation}
d_\varrho|_{z = H_0} = 0, \quad d_\varrho|_{z = 0} = 0
\end{equation}
which can deform the cylinder in an hourglass-like shape under stretching load and bulge outward like a drum under the compression load, exhibiting distinct deformation patterns.

For the problem above with complex boundary conditions, the detailed settings of the CAG method are reasonably determined. The axial stretch ration $\xi$ is set as input, and displacements $d_x$, $d_y$, and $d_z$ in the Cartesian coordinate system, derived from $d_\varrho$, $d_\vartheta$, and $d_z$, as outputs. The input range spans from 0.52 to 1.28. For the adaptive sampling process, the settings of $K=2$, $Q_\text{min}=10$, and $M^0 = 21$ lead to a final training sample size of $M=26$. The physical field dimension is $N=3,317$ according to the number of nodes in the high-fidelity FEM solver, and the reduced dimension is $R=40$. The prediction includes 8 representative cases uniformly selected between 0.5 and 1.9. Additionally, the GPR with uniform sampling is employed for comparative purposes.
\begin{figure}[tbph]
    \begin{centering}
    \includegraphics[width=14.5cm]{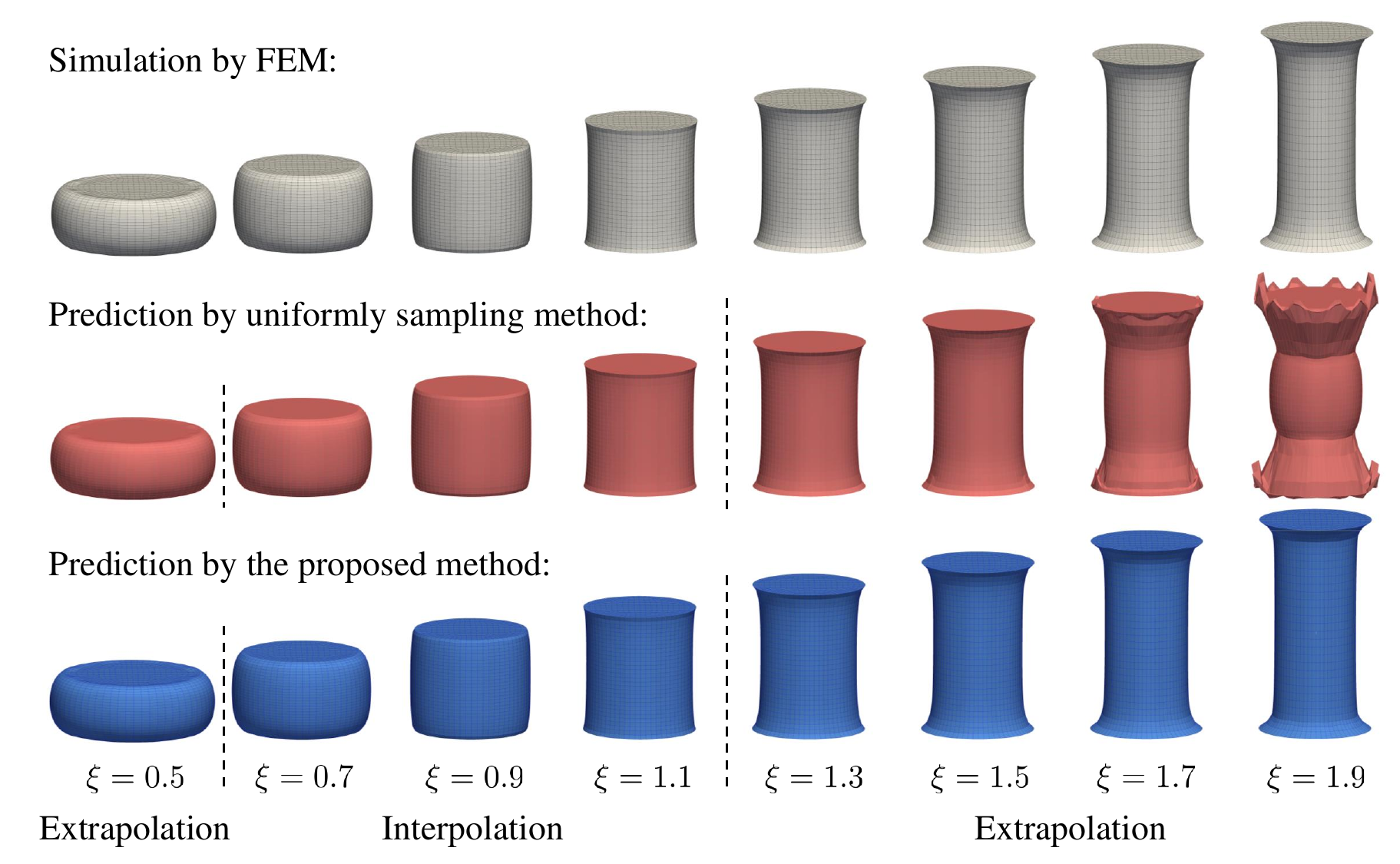}
    \par\end{centering}
    \caption{A side-by-side comparison of results from FEM, the GPR with uniform sampling, and the proposed CAG method for the hyperelastic cylinder under various stretch ratios.}
\label{fig_hyper_multi}
\end{figure}

\autoref{fig_hyper_multi} compares the configurations predicted by FEM, the uniform sampling method, and the proposed method across eight testing cases. Notably, five of the eight inputs fall outside the training range of $[0.52,1.28]$, labeled as extrapolation in the figure, while the remaining three cases are interpolation.
For $\xi = 1.7$ and $\xi=1.9$, the uniform sampling method exhibits poor performance, resulting in abnormal shapes. Specifically, the drum-like compression pattern interferes with the hourglass-like stretching pattern, leading to bulges in the middle. 
In contrast, the proposed method effectively mitigates these issues by automatically partitioning the training set into stretching and compression patterns. This partitioning helps minimize compression training sets' impact on stretching predictions.
Quantitatively, the extrapolation means square error (MSE) for $d_x$, $d_y$, and $d_z$ using the uniform sampling method is $3.22\times 10^{-4}$, $3.00\times 10^{-4}$ and $4.60\times 10^{-3}$, respectively. In contrast, the proposed CAG method achieves MSE values of $3.65\times 10^{-5}$, $3.70\times 10^{-5}$ and $1.77\times 10^{-5}$, respectively, indicating a one-order-of-magnitude improvement in accuracy for $d_x$ and $d_y$ and a two-order-of-magnitude improvement for $d_z$. It is evident that the proposed CAG method exhibits improved generalization ability thanks to pattern partitioning. Detailed prediction errors are listed in \autoref{tab_hyper_detail} in Appendix A. 

\subsection{Elastoplastic T-shaped structure under torsion}
\label{subsectwist}

The fourth example is an elastoplastic T-shaped structure under torsion, involving both material and geometric nonlinearities. 
In this problem, the impact of the cluster number $K$ on the performance of the proposed method is discussed.

\begin{figure}[tbph]
    \begin{centering}
    \includegraphics[width=5cm]{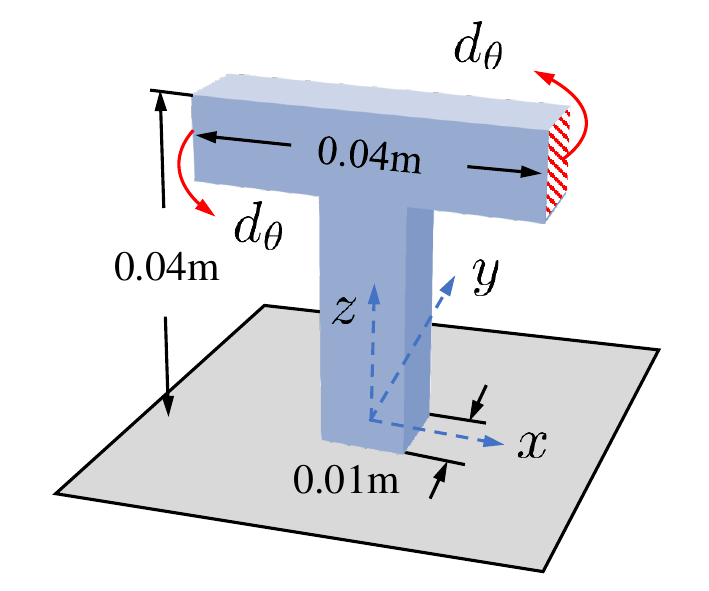}
    \par\end{centering}
    \caption{Schematic diagram of the T-shaped structure under torsion.}
\label{fig_twist_sk}
\end{figure}

The T-shaped structure's shape and dimensions are illustrated in \autoref{fig_twist_sk}. The bottom end of the structure is fixed, and the two ends of the top beam are under circumferential displacement denoted as $d_\theta$. A positive $d_\theta$ corresponds to the counterclockwise direction in the top view, while a negative value indicates the clockwise direction. The elastoplastic material takes into account kinematic hardening using the $J_2$ flow criterion, with an elastic modulus of 200 GPa, a Poisson's ratio of 0.3, a yield stress of 235 MPa, and a plastic tangent modulus of 6.1 GPa.

In this problem, the $d_\theta$ is set as input ranging from -22 mm to 22 mm, and four variables are set as output, including $d_x$, $d_y$, $d_z$ displacements in the Cartesian coordinate system, and the von-Mises strain $\varepsilon_{\text{Mises}}$.
Four models are compared to evaluate the impact of cluster number $K$, including the proposed method with 1 cluster, 2 clusters, 3 clusters, and 4 clusters. Pattern partitioning is achieved using $\varepsilon_{\text{Mises}}$. To ensure a fair comparison, all models utilize the same training sample size of 22. Specifically, for $K=1$, the model degenerates to GPR with uniform sampling using $M^0 = 22$. For $K=2$, $M^0 = 12$ and $Q_\text{min}=8$ are utilized. For $K=3$, $M^0 = 12$ and $Q_\text{min}=6$ are employed. For $K=4$, $M^0 = 10$ and $Q_\text{min}=4$ are employed. Both $M^0$ and $Q_\text{min}$ are chosen so that the final sample size $M$ remains at 22 for all cases.
The physical field dimension is $N=1,332$, which corresponds to the number of nodes in the FEM solver, and the reduced dimension is set to $R=40$. 
The prediction consists of 10 representative cases uniformly distributed between -21.5 mm and 21.5 mm, encompassing both clockwise and counterclockwise torsion.

\begin{figure}[tbph]
    \begin{centering}
    \includegraphics[width=14cm]{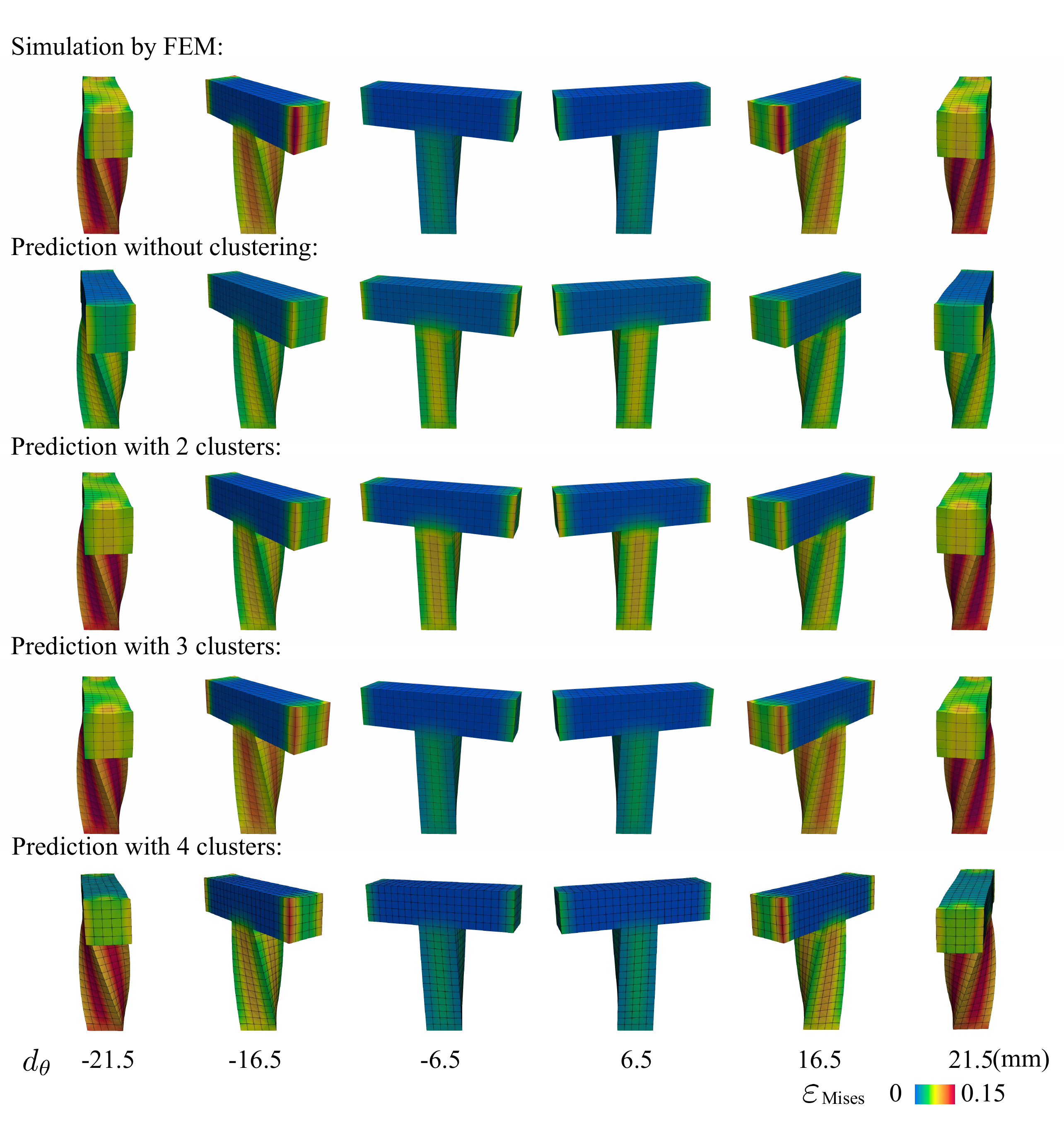}
    \par\end{centering}
    \caption{Deformation and strain of the T-shaped structure under torsion: FEM results and predictions from the proposed CAG method with different clusters.}
\label{fig_twist_comp}
\end{figure}

Six representative cases from the 10 cases are first studied.
As shown in \autoref{fig_twist_comp}, the predicted equivalent strain and deformed configurations from the three models are compared.
It is clear that the accuracy of equivalent strain predictions varies significantly. 
The model with $K=1$ underestimates the strain in the supporting column for cases with $d_\theta$ values of $\pm$21.5 mm, as well as $\pm$16.5 mm, and overestimates the strain for the case with $d_\theta=\pm$6.5 mm. 
In contrast, the model with $K=2$ demonstrates improved performance in predicting high strain regions for the case with $d_\theta=\pm$21.5 mm, but makes few improvements for the remaining four cases. As $K=3$, the model yields all predictions for the six cases aligned closely with the FEM results. As $K=4$, the predictions are similar to those observed with $K=3$, except for a minor reduction in the upper beam for the case with $d_\theta=\pm$21.5 mm.

The reason for different prediction performances can be attributed to the distribution of cases and their corresponding strain patterns. 
With $K=1$, the strain patterns of all 22 training samples are averaged, exhibiting near-zero strain in the middle of the upper beam, small strain in the two ends of the upper beam, and in the column support. As a result, all six unseen cases are assumed to have a consistent strain pattern above, which does not align with the fact. 
As $K$ increases to $2$, there are two clusters, Cluster A and Cluster B. The samples with the value of $d_\theta$ in the range of $[-22,-19.5)\cup (19.5,22]$ belong to Cluster A, while Cluster B for the $d_\theta$ within $[-19.5,19.5]$. 
As a consequence, the cases with $\pm$21.5 mm in Cluster A are individually predicted, exhibiting high strain in the supporting column. The remaining four cases in Cluster B maintain a similar pattern to that of $K=1$. 
As $K=3$, there are three clusters: Cluster A for $d_\theta$ within $[-22,-20.25)\cup (20.25,22]$, Cluster B for $d_\theta$ within $[-20.25,-13)\cup (13,20.25]$, and Cluster C for $d_\theta$ within $[-13,13]$. The $\pm$16.5 mm cases fall into Cluster B, better highlighting the elevated strain in the supporting column, while the $\pm$6.5 mm cases in Cluster C accurately depict the near-zero strain pattern.
For $K=4$, the deciphered clusters are: Cluster A for $d_\theta$ within $[-22,-21.5)\cup (21.5,22]$, Cluster B for $d_\theta$ within $[-21.5,-18.5)\cup (18.5,21.5]$, Cluster C for $d_\theta$ within $[-18.5,-11)\cup (11,18.5]$, and Cluster D for $d_\theta$ within $[-11,11]$. Its prediction performance is similar to that of $K=3$. 

For quantitative comparison, the mean square errors across all ten prediction scenarios are detailed in \autoref{tab_twist_comp}, indicating a statistical measure of predictive accuracy. A more detailed comparison of the prediction errors for each scenario is provided in \autoref{tab_twist_comp_detail} in Appendix A. 
\begin{table}[]
\small
\caption{Comparative evaluation of mean square error for strain and displacement among different clusters}
\centering
\begin{tabular}{ccccc}
    \toprule
    Cluster number & MSE for $\varepsilon_{\text{Mises}}$ &MSE for $d_x$ & MSE for $d_y$ & MSE for $d_z$  \\
    \midrule
    $K=1$  & $1.06\times 10^{-3}$ & $2.78\times 10^{-8}$ & $1.95\times 10^{-8}$ & $4.17\times 10^{-9}$ \\
    $K=2$  & $3.60\times 10^{-4}$ & $3.03\times 10^{-8}$ & $3.84\times 10^{-9}$ & $1.53\times 10^{-9}$\\
    $K=3$  & $1.19\times 10^{-4}$ & $5.11\times 10^{-10}$ & $1.52\times 10^{-9}$ & $3.20\times 10^{-10}$\\
    $K=4$  & $2.72\times 10^{-4}$ & $1.13\times 10^{-7}$ & $1.80\times 10^{-9}$ & $6.81\times 10^{-9}$\\
    \bottomrule
\label{tab_twist_comp}
\end{tabular}
\end{table}
The comparison demonstrates a clear improvement in predictive precision achieved through clustering. This highlights the method's ability to learn from subtle similarities among patterns and to make accurate predictions, especially for complex problems with significant variations in physical field patterns. Specifically, despite fluctuations observed in individual cases (detailed in Appendix A), the overall MSE for strain and displacement at $K=2$ is reduced by one to two orders of magnitude compared to $K=1$. At $K=3$, the overall MSE decreases, whereas at $K=4$, it slightly increases compared to $K=3$. The reasons for this trend are further discussed below. 

Theoretically, the prediction accuracy improves with an increasing number of clusters $K$ as long as each cluster has a sufficient sample size, $\hat{M}_k$. This is because a higher cluster count allows for more detailed discrimination of response patterns.
However, in practical settings, computational resources for training samples are often constrained, limiting the total number of samples $M$ that can be prepared (as exemplified in the T-shaped structure case, where $M=22$ is assumed). Under such constraints, increasing $K$ necessarily reduces the sample size, $\hat{M}_k$, within each cluster, which can compromise the prediction accuracy.
Therefore, the selection of $K$ requires a balance between the benefits of finer pattern division and the accuracy losses resulting from reduced sample sizes within clusters. 
For example, for the T-shaped structure problem, when $K=4$, a specific prediction scenario for $d_\theta=-21.5$ is conducted based on only four samples. Its prediction error is higher than the counterpart with $K=3$, and the same scenario is predicted with six samples. 
Consequently, the fundamental principle in selecting $K$ in practical applications is to assess the available total sample size and the complexity of structural responses. Ensuring at least 6 to 8 samples per cluster is advisable, and a recommended $K$ value of 2 or 3 is suitable for most practical scenarios based on tests conducted in this study.

\subsection{Circular ring under high-speed impact}

The fifth example addresses the dynamic structural response, incorporating strong material, geometric, and contact nonlinearities. In particular, it accounts for the deformation of an elastoplastic circular ring under high-speed impact, which has been experimentally investigated by Xu et al. \cite{xu_collision_2015}.

\begin{figure}[tbph]
    \begin{centering}
    \includegraphics[width=4.5cm]{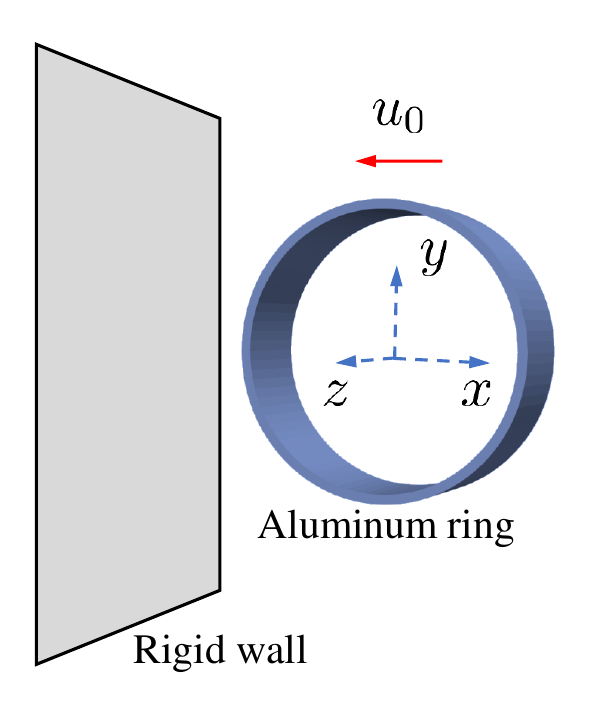}
    \par\end{centering}
    \caption{Schematic diagram of the collision of an aluminum ring against a rigid wall, where $u_0$ denotes the initial velocity.}
\label{fig_ring_sk}
\end{figure}

\autoref{fig_ring_sk} illustrates the geometry of the circular ring colliding with a rigid wall at an initial velocity of $u_0$. According to the experimental data, the ring has a diameter of 28.5 mm, a thickness of 0.91 mm, and a width of 9.1 mm. The material is the elastoplastic aluminum alloy, with a density of $2270 \;\rm kg/m^3$, a Young's modulus of $70 \;\rm GPa$, a Poisson’s ratio of $0.3$, a yield strength of $300 \;\rm MPa$, and a plastic tangent modulus of $1 \;\rm GPa$.

For this example, we employed $u_0$ as input, varying from 22 m/s to 126 m/s, and use displacements $d_x$, $d_y$, and $d_z$ as outputs. To eliminate the influence of rigid body motion, the displacements $d_x$, $d_y$, and $d_z$ are considered as relative displacements with respect to the centroid of the deformed ring. The sample generation process uses parameters of $K=2$, $M^0 = 8$ and $Q_\text{min}=7$, resulting in a final sample size of $M=14$. The explicit dynamics FEM software LS-DYNA is employed as the solver $\mathcal S$, utilizing the Hughes-Liu shell element \citep{hughes1981nonlinear} with 4 Gaussian integration points across the thickness. 
The physical field dimension determined by the number of nodes is $N=900$, and the reduced dimension is determined as $R=40$. 
The testing set replicates the impact velocities used in the experiments, with a sample count of impact velocities identical to those used in the experiments, with unseen cases of $M^*=7$, and $u_0$ values ranging from 23.8 m/s to 125 m/s. The GPR with uniform sampling is conducted as well for comparison, and evaluation is performed by comparing the mean squared error to the FEM simulation results.

\begin{figure}[tbph]
    \begin{centering}
    \includegraphics[width=13cm]{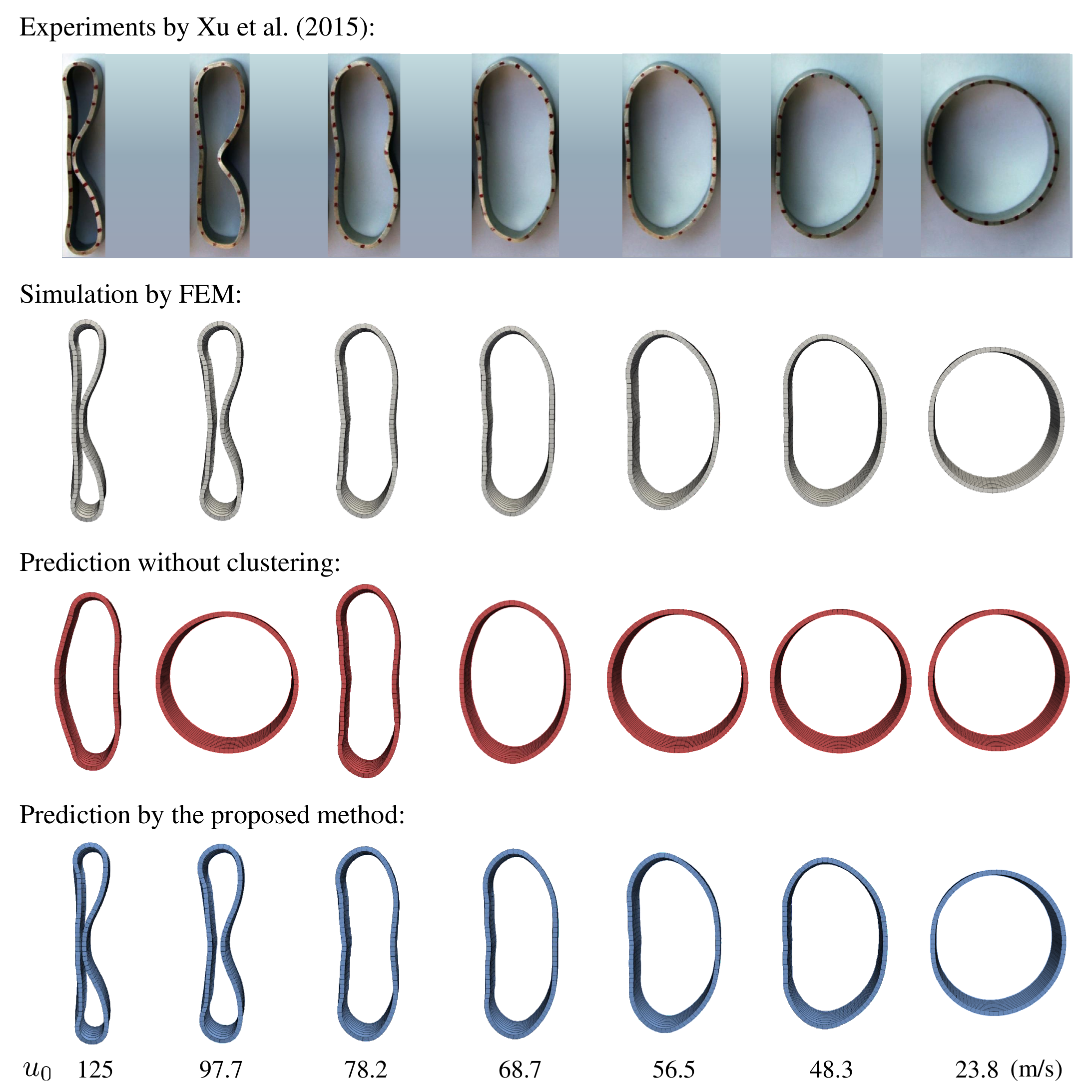}
    \par\end{centering}
    \caption{Aluminum ring configurations after impact: Comparisons of experiments, FEM results, predictions using GPR with uniform sampling, and predictions by the proposed CAG method.}
\label{fig_ring_comp}
\end{figure}

\autoref{fig_ring_comp} compares the experimental data, FEM simulation results, and predictions using the GPR with uniform sampling and those achieved by the proposed method. It is evident that the FEM results align closely with the experimental data, and the predictions generated by the proposed method align well with the FEM results. In contrast, the uniform sampling method leads to inaccurate predictions, as distinct deformation patterns interfere with each other.

\autoref{tab_ring_comp} lists the mean square errors in displacements across all seven prediction cases, and \autoref{tab_ring_detail} details the specific error for each case. The results indicate that the proposed method exhibits significantly lower MSE values in both $d_x$ and $d_y$ compared to the counterparts in GPR with uniform sampling, with a difference of three orders of magnitude. The MSE values for $d_z$ are negligible because $d_z$ is approximately zero throughout the entire ring. 
\begin{table}[]
\small
\caption{Comparison of mean square error in displacements for the ring under high-speed impact.}
\centering
\begin{tabular}{cccc}
\toprule
Method   &MSE for $d_x$ & MSE for $d_y$ & MSE for $d_z$  \\
\midrule
Uniformly sampling & $1.21\times 10^{-5}$ &  $1.92\times 10^{-6}$ &   $6.54 \times 10^{-10}$  \\
Proposed method  &  $1.71\times 10^{-8}$ &  $1.00\times 10^{-9}$ &   $4.54 \times 10^{-10}$ \\
\bottomrule
\label{tab_ring_comp}
\end{tabular}
\end{table}

\subsection{Brittle plate with dynamic crack branching}\label{Plate}

The final example presents a complex fracture problem.
In this problem, a brittle pre-cracked plate subject to boundary tension presents the dynamic crack branching, as reported in \cite{belytschko2003dynamic,kakouris2019phase,zeng2023explicit}. 
The crack profiles are chosen to validate the performance of the proposed method. 

\begin{figure}[tbph]
    \begin{centering}
    \includegraphics[width=6.5cm]{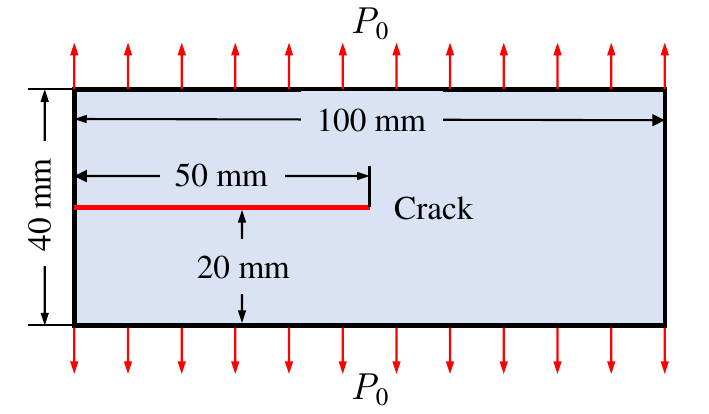}
    \par\end{centering}
    \caption{Schematic diagram of a brittle pre-cracked plate under transient tensile loads.}
\label{fig_crack_sch}
\end{figure}

The dimensions of the pre-cracked plate are illustrated in \autoref{fig_crack_sch}, assuming plain strain conditions. A constant tensile load, labeled as $P_0$ is uniformly distributed on both top and bottom edges of the plate. Material parameters include a density of $2450 \;\rm kg/m^3$, a Young's modulus of 32 GPa, a Poisson's ratio of 0.2, and an energy release rate of $3.0\;\rm J/m^2$. 

In this example, we employ the tensile load $P_0$ as input, ranging from 0.7 MPa to 0.99 MPa, and the damage indicator $\Tilde{ \Theta }$ as output. $\Tilde{ \Theta }$ is calculated by using the following equation.
\begin{equation}
\Tilde{ \Theta } = \dfrac{1}{1+\exp \left( -15( \Theta -0.5)  \right) }
\end{equation}
where $\Theta$ represents the damage obtained from the phase field finite element method (PFFEM) \citep{borden2012phase} solver $\mathcal S$. The Sigmoid function is utilized to map the damage to the range of $[0,1]$, ensuring a sharper representation of the crack.  
In the PFFEM model, an element size of 0.125 mm, a critical length of 1 mm, a viscous dissipation parameter of 0.01, and a fixed time step size of $2 \times 10^{-8}$ s are adopted. The simulations were conducted on a supercomputer with 56-core Intel Xeon E5-2680V4 CPUs. 
Although the total number of element nodes is 257,121, the crack can be effectively represented using a smaller dimension. Therefore, a linear interpolation is applied to transform the damage into a 20,000-dimension array, resulting in $N=20,000$.
The reduced dimension is determined as $R=40$. The sample generation process uses parameters of $K=3$, $M^0 = 8$, and $Q_\text{min}=4$, resulting in a final sample size of $M=15$. 
$M^*=5$ cases with representative values of representative $P_0$ being 0.74, 0.77, 0.83, 0.88, and 0.94 MPa, resepctively, are conducted. Additionally, the GPR without clustering is employed for comparison purposes. 

\begin{figure}[tbph]
    \begin{centering}
    \includegraphics[width=15.5cm]{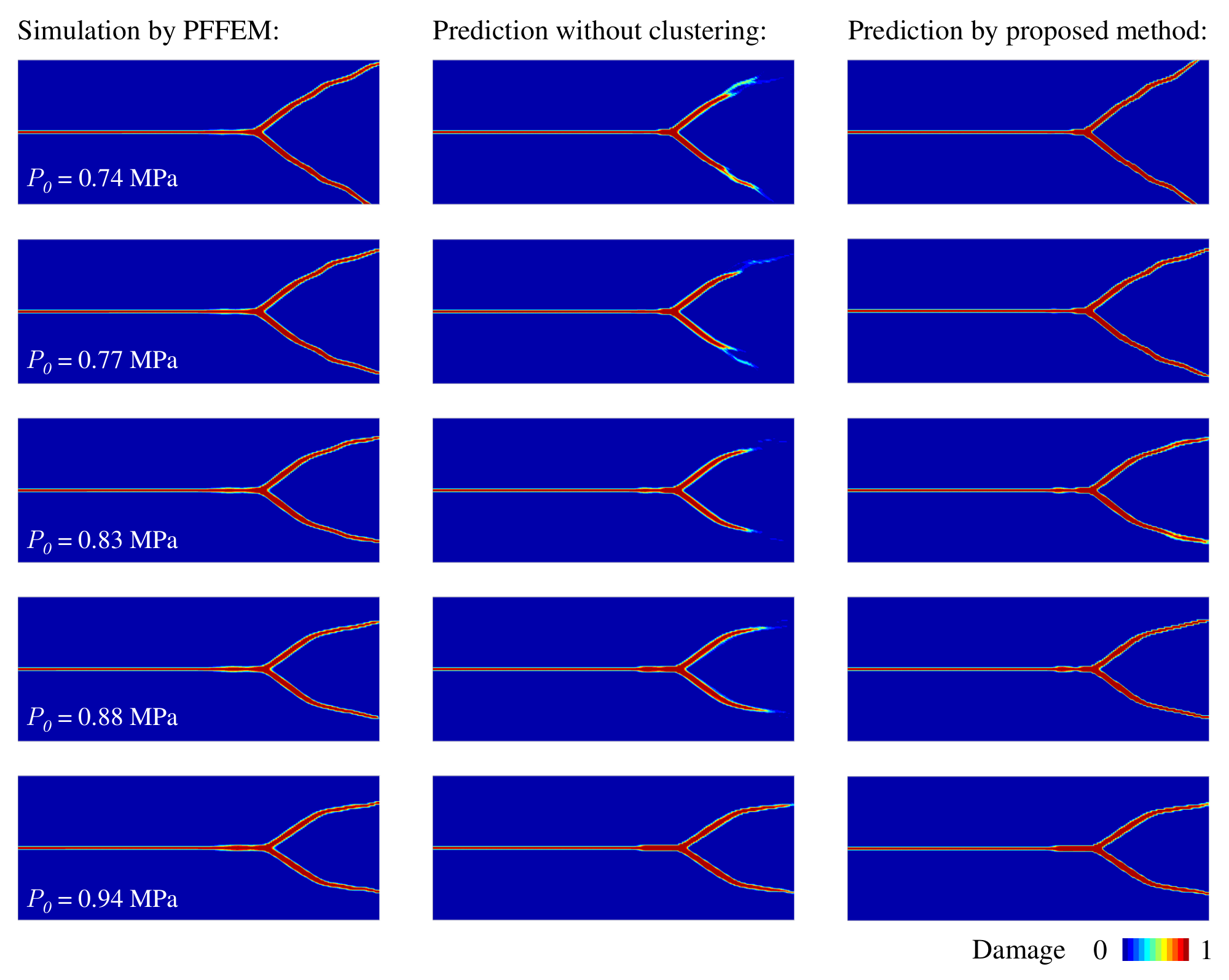}
    \par\end{centering}
    \caption{Comparison of morphology of the branching cracks between the numerical simulation results and predictions from the proposed CAG method and the GPR without clustering.}
\label{fig_frac_comp}
\end{figure}

\autoref{fig_frac_comp} compares the PFFEM results, the prediction by the GPR without clustering, and the prediction by the proposed CAG method with 3 clusters. It is evident that varying loads $P_0$ lead to distinct branched crack paths near the right edge, resulting in diverse damage patterns. This complexity brings challenges for accurate predictions for the GPR without clustering, where crack tips of branches are influenced by other patterns, leading to their merging and disappearance near the right edge. This limits the GPR's ability to predict the crack extending through the entire plate. However, the proposed method with three clusters capturing the feature offers more precise predictions. In terms of quantitative evaluation, we calculate the relative error of total damage by summing $\Tilde{ \Theta }$ over the entire domain. Without clustering, the GPR predicts five scenarios with an average error of 15.05\%. In contrast, the proposed CAG method with three clusters demonstrates an error of only 3.54\%. The details of these errors are outlined in \autoref{tab_crack_detail}.

\subsection{Evaluation of prediction efficiency}

For quantitative comparison of the efficiency of the proposed CAG method, the computational costs for the four problems from Sec. \ref{CylinderProblem} to Sec. \ref{Plate} are listed in \autoref{tab_time}. 
All computations except for the crack branching simulation, which utilized 56 cores on a supercomputer, are conducted on a desktop computer equipped with an Intel i7-13700F CPU. 
Notably, the online stage of the proposed method consistently operates within 1 second, in favor of the need for real-time evaluation. In comparison to high-fidelity simulations, the proposed method has significantly reduced computational costs by three to six orders of magnitude.
It is important to point out that the proposed method excels at predicting many cases due to the utilization of matrix calculations in Eqs.~\autoref{eq_predict} and \autoref{eq_restoration}. 
Consequently, the time required to process numerous cases does not increase linearly with the number of cases. For instance, predicting a single case considering the cylinder under stretching/compression requires 0.105 seconds, while predicting 100 samples only takes 0.111 seconds, indicating a minimal increase in time consumption. In contrast, high-fidelity simulations for 100 cases typically consume 100 times more time than for a single case.
\begin{table}[]
\small
\caption{Computational costs of the predictions of different cases (unit in second)}
\centering
\begin{tabular}{ccccc}
    \toprule
    Case  & High-fidelity simulation  &   Proposed CAG method  \\
    \midrule
    Cylinder under stretching/compression &    40      &   0.105    \\
    T-shaped structure under torsion  &    16     &    0.244    \\
    Ring under high-speed impact  &   434    &     0.254   \\
    Brittle plate with crack branching  &  29,580  &    0.155 \\
    \bottomrule
\label{tab_time}
\end{tabular}
\end{table}

\section{Conclusion}
\label{secCon}

In this study, a clustering adaptive Gaussian process regression (CAG) method is proposed to tackle intricate solid mechanics challenges exhibiting diverse behavior patterns.
It consists of the clustering-enhanced adaptive sample generation, the training of multi-pattern Gaussian process regressor, dimensionality reduction and restoration techniques, pre-prediction classification, and online prediction using multi-pattern Gaussian process regressor, where the demand-driven sample allocation and the divide-and-conquer strategy are fulfilled for the offline and online stages, respectively. 

Six problems, considering nonlinearities in material, geometry, and boundary conditions, demonstrate the method's predictive capabilities. Notably, CAG exhibits exceptional accuracy, especially with small sample sizes, reducing errors by $1 \sim 3$ orders of magnitude compared to traditional Gaussian process regression models with uniformly distributed samples. 
Remarkably, the proposed CAG method only requires around 20 training samples to obtain accurate predictions for the examples studied in this work without significant computational costs in sample construction. Meanwhile, compared to high-fidelity simulations, it significantly reduces online time costs by $3\sim 6 $ orders of magnitude. 

By harnessing the power of demand-driven sample allocation and the divide-and-conquer strategy, the proposed CAG method demonstrates a high level of efficiency in analyzing nonlinear solid mechanics problems. This efficiency, in turn, accelerates the process of revealing the underlying mechanisms in their response patterns.   
In complex solid mechanics problems, employing unsupervised methods to ascertain the number of response patterns rather than relying on a predefined value offers a promising avenue. 
Therefore, future work to improve this method will focus on adaptively determining the optimal number of structural response patterns.
On this basis, identifying structural response patterns, gaining a deeper understanding of the critical conditions for pattern shifts, and even exploring hitherto unrecognized patterns constitute valuable areas for further exploration.
Looking ahead, we envision the CAG method being tailored to a wide range of applications, including online structural health monitoring, digital twins, and other scenarios that demand real-time prediction of structural responses with rich details. This potential for diverse applications is a testament to the versatility and robustness of the proposed method.   

\section*{Declaration of competing interest}

The authors declare that they have no known competing financial interests or personal relationships that could have appeared to influence the work reported in this paper.

\section*{Data availability}

Data will be made available on request.

\section*{Acknowledgements}

The authors acknowledge the support from the  National Key R\&D Program of China under Grant No. 2023YFB3309100 and 2023YFB3309105, the National Natural Science Foundation of China under Grant No. 52105287, the Fundamental Research Funds for the Central Universities, and the Beijing Institute of Technology Research Fund Program for Young Scholars (XSQD-202223002).

\appendix
\section{Prediction error for specific prediction cases}
\label{AppendixA}
\setcounter{table}{0}

Detailed prediction errors of traditional GPR and the proposed CAG method are listed here for the following scenarios: the cylinder under stretching/compression, the T-shaped structure under torsion, the ring under high-speed impact, and the crack branching in the brittle plate.  

\begin{table}[]
\small
\caption{Mean square error of displacement for cases in cylinder under stretching/compression} 
\centering
\begin{tabular}{ccccccc}
    \toprule
    &\multicolumn{3}{c}{GPR} &\multicolumn{3}{c}{CAG}\\
  $\xi$  &$d_x$  &$d_y$   &$d_z$    &$d_x$    &$d_y$   &$d_z$     \\
     \midrule
0.5 &$1.72\times 10^{-8}$   &$4.45\times 10^{-9}$   &$9.23\times 10^{-9}$
&$2.02\times 10^{-8}$  &$5.54\times 10^{-9}$  &$1.24\times 10^{-8}$\\

0.7 &$ 2.14\times 10^{-11}$   &$ 1.62\times 10^{-11}$   &$ 2.11\times 10^{-11}$ 
&$4.23\times 10^{-11}$  &$3.28\times 10^{-11}$  &$4.47\times 10^{-11}$\\

0.9 &$ 2.36\times 10^{-10}$   &$ 2.29\times 10^{-10}$   &$ 1.68\times 10^{-10}$ 
&$1.73\times 10^{-10}$  &$1.69\times 10^{-10}$  &$1.47\times 10^{-10}$\\

1.1 &$ 3.46\times 10^{-12}$   &$ 3.95\times 10^{-12}$   &$ 1.01\times 10^{-11}$ 
&$9.12\times 10^{-12}$  &$9.42\times 10^{-12}$  &$1.57\times 10^{-11}$\\

1.3 &$ 1.54\times 10^{-10}$    &$ 1.62\times 10^{-10}$    &$ 3.00\times 10^{-10}$ 
&$3.73\times 10^{-11}$  &$3.74\times 10^{-11}$  &$7.37\times 10^{-12}$\\

1.5 &$ 2.36\times 10^{-6}$    &$ 2.59\times 10^{-6}$    &$ 1.01\times 10^{-5}$ 
&$3.17\times 10^{-7}$  &$3.20\times 10^{-7}$  &$8.30\times 10^{-8}$\\

1.7 &$ 1.37\times 10^{-4}$    &$ 1.34\times 10^{-4}$    &$ 1.05\times 10^{-3}$ 
&$1.30\times 10^{-5}$  &$1.32\times 10^{-5}$  &$4.85\times 10^{-6}$\\

1.9&$ 1.15\times 10^{-3}$    &$ 1.06\times 10^{-3}$    &$ 1.74\times 10^{-2}$ 
&$1.33\times 10^{-4}$  &$1.35\times 10^{-4}$  &$6.58\times 10^{-5}$\\
     \bottomrule
\label{tab_hyper_detail}
\end{tabular}
\end{table}

\begin{table}[]
\small
\caption{Mean square error of strain and displacement for the cases considering T-shaped structure under torsion. Cases with positive circumferential displacement $d_\theta$ are not included in this table, since the structural deformations are symmetrical to the ones presented, and the prediction errors are the same. } 
\centering
\begin{tabular}{ccccccc}
    \toprule
      &$d_\theta$ (mm)  & -21.5 & -16.5  & -0.0115  & -0.0065 &  -0.0015     \\
    \midrule
    \multirow{4}{*}{$K$=1} &$\varepsilon_{\text{Mises}}$ &$ 3.36\times10^{-3}$ &$ 2.94\times 10^{-4}$ &$ 8.24\times 10^{-5}$ &$  4.14\times 10^{-4}$ &$ 1.15\times10^{-3}$ \\
                        &$d_x$ &$ 6.20\times 10^{-8 }$ &$ 5.88\times 10^{-8 }$ &$ 5.99\times 10^{-9  }$ &$  3.31\times 10^{-9 }$ &$  8.63\times 10^{-9 }$ \\
                        &$d_y$ &$ 4.16\times 10^{-8  }$ &$  2.88\times 10^{-8  }$ &$ 8.81\times 10^{-9  }$ &$ 2.60\times 10^{-9 }$ &$ 1.56\times 10^{-8 }$ \\
                        &$d_z$ &$ 1.93\times 10^{-8 }$ &$  1.27\times 10^{-9  }$ &$ 1.66\times 10^{-10 }$ &$ 3.29\times 10^{-11 }$ &$  3.81\times 10^{-11 }$ \\ 
 \midrule
    \multirow{4}{*}{$K$=2} &$\varepsilon_{\text{Mises}}$ &$ 6.27\times 10^{-6}$ &$ 2.57\times10^{-4 }$ &$ 3.16\times 10^{-5 }$ &$  3.76\times 10^{-4}$ &$ 1.13\times10^{-3 }$ \\
                        &$d_x$  &$ 1.48\times 10^{-7 }$ &$1.45\times 10^{-9 }$ &$8.65\times 10^{-10  }$ &$ 5.21\times 10^{-10 }$ &$  1.93\times 10^{-10 }$ \\
                        &$d_y$ &$ 1.26\times 10^{-9  }$ &$ 5.76\times 10^{-9  }$ &$  5.71\times 10^{-9  }$ &$ 7.10\times 10^{-10 }$ &$ 5.78\times 10^{-9  }$ \\
                        &$d_z$ &$ 7.62\times 10^{-9 }$ &$ 1.01\times 10^{-12  }$ &$ 8.23\times 10^{-13 }$ &$ 4.27\times 10^{-13 }$ &$ 4.35\times 10^{-13 }$ \\ 
 \midrule
    \multirow{4}{*}{$K$=3}&$\varepsilon_{\text{Mises}}$ &$ 6.75\times 10^{-5}$ &$  4.06\times 10^{-5  }$ &$ 1.89\times 10^{-4  }$ &$ 8.02\times 10^{-6 }$ &$ 2.91\times 10^{-4 }$ \\
                        & $d_x$  &$ 3.75\times 10^{-10 }$ &$ 1.05\times 10^{-9  }$ &$ 2.88\times 10^{-11  }$ &$ 5.26\times 10^{-10 }$ &$ 5.75\times 10^{-10 }$ \\
                        &$d_y$  &$ 1.68\times 10^{-9  }$ &$ 5.86\times 10^{-9  }$ &$ 2.54\times 10^{-12  }$ &$ 3.27\times 10^{-11 }$ &$ 3.84\times 10^{-11  }$ \\
                        & $d_z$  &$ 1.60\times 10^{-9 }$ &$ 7.73\times 10^{-13  }$ &$  9.85\times 10^{-15 }$ &$ 1.39\times 10^{-13 }$ &$ 4.85\times 10^{-13 }$ \\
 \midrule
    \multirow{4}{*}{$K$=4} &$\varepsilon_{\text{Mises}}$  &$ 9.03\times 10^{-4}$ &$   1.69\times 10^{-4 }$ &$ 4.06\times 10^{-5  }$ &$ 2.35\times 10^{-6   }$ &$ 2.45\times 10^{-4 }$ \\
                        & $d_x$  &$  5.64\times 10^{-7 }$ &$ 5.49\times 10^{-11 }$ &$3.02\times 10^{-10  }$ &$2.00\times 10^{-10   }$ &$  1.09\times 10^{-10 }$ \\
                        &$d_y$  &$  8.58\times 10^{-9  }$ &$ 7.81\times 10^{-12 }$ &$3.99\times 10^{-10  }$ &$1.17\times 10^{-11   }$ &$  9.73\times 10^{-12 }$ \\
                        & $d_z$  &$  3.40\times 10^{-8 }$ &$ 8.72\times 10^{-15 }$ &$9.96\times 10^{-14  }$ &$1.91\times 10^{-13   }$ &$ 3.67\times 10^{-13 }$ \\
     \bottomrule
\label{tab_twist_comp_detail}
\end{tabular}
\end{table}

\begin{table}[]
\small
\caption{Mean square error of displacement for cases in ring under high-speed impact. } 
\centering
\begin{tabular}{cccccccccc}
    \toprule
    &\multicolumn{3}{c}{GPR} &\multicolumn{3}{c}{CAG}\\
  $u_0\;$(m/s)  &$d_x$  &$d_y$ &$d_z$  &$d_x$  &$d_y$ &$d_z$   \\
   \midrule
   125  &$2.21 \times 10^{-3}$   &$9.34 \times 10^{-1}$   & 10.93
    &$3.06\!\times\! 10^{-3}$  &$2.66\!\times\! 10^{-3}$  &$4.74\!\times\! 10^{-2}$   \\

 97.7 &$1.58\times 10^{-3}$   &6.10   & 56.03  &$ 9.53\times 10^{-5}$  &$ 3.63\times 10^{-3}$  &$ 6.79\times 10^{-2}$   \\

 78.2 &$2.71\times 10^{-6}$  &$2.57\times 10^{-3}$  &$1.54\times 10^{-2}$
  &$ 5.96\times 10^{-6}$  &$ 1.03\times 10^{-4}$  &$ 1.70\times 10^{-4}$   \\

 68.7 &$2.18\times 10^{-4}$  &1.57   &5.52  
  &$ 7.64\times 10^{-6}$  &$ 1.86\times 10^{-4}$  &$ 1.26\times 10^{-3}$   \\

 56.5 &$3.48\times 10^{-4}$  &2.94   &7.93  
  &$ 4.06\times 10^{-6}$  &$ 9.75\times 10^{-5}$  &$ 1.20\times 10^{-3}$   \\

 48.3 &$2.10\times 10^{-4}$  &1.85   &4.18  
  &$ 4.25\times 10^{-6}$  &$ 5.54\times 10^{-4}$  &$ 1.61\times 10^{-3}$   \\

 23.8 &$6.21\times 10^{-6}$  &$5.47\times 10^{-2}$  &$7.27\times 10^{-2}$
  &$ 4.50\times 10^{-6}$  &$ 1.11\times 10^{-4}$ &$ 2.75\times 10^{-4}$  \\
     \bottomrule
\label{tab_ring_detail}
\end{tabular}
\end{table}

\begin{table}[]
\small
\caption{Relative errors of the total damage for cases considering crack branching in brittle plate.} 
\centering
\begin{tabular}{cccccc}
    \toprule
    $P_0\;$(MPa)  &  0.74  &  0.77  & 0.83   & 0.88  &  0.94 \\
     \midrule
GPR  &22.70\%  	&23.64\% 	&18.16\% 	&7.13\%    &2.49\%  \\
CAG    &7.80\% 	&2.98\%	&2.53\% &3.29\% 	&0.87\%      \\
     \bottomrule
\label{tab_crack_detail}
\end{tabular}
\end{table}

\bibliographystyle{elsarticle-num}
\bibliography{main}
\biboptions{numbers,sort&compress}

\end{document}